\documentclass[5p,times]{elsarticle}
\usepackage{hyperref}

\usepackage{amsmath,amssymb,amsfonts}
\usepackage{graphicx}
\usepackage{textcomp}
\usepackage[caption=false]{subfig}
\usepackage{xcolor}
\usepackage{caption}
\usepackage{comment}
\usepackage{todonotes}
\usepackage{bbm}
\usepackage[linesnumbered,ruled,vlined]{algorithm2e}

\SetCommentSty{mycommfont}
\usepackage{wrapfig}

\begin{document}

\begin{frontmatter}


\title{Explaining the Deep Natural Language Processing by Mining Textual Interpretable Features}

\author{Francesco Ventura\corref{mycorrespondingauthor}}
\ead{francesco.ventura@polito.it}

\author{Salvatore Greco}
\ead{salvatore_greco@polito.it}

\author{Daniele Apiletti}
\ead{daniele.apiletti@polito.it}

\author{Tania Cerquitelli}
\ead{tania.cerquitelli@polito.it}

\address{Department
of Control and Computer Engineering, Politecnico di Torino, Turin, Italy}
\cortext[mycorrespondingauthor]{Corresponding author}

\newcommand{\XXX}{{\sc T-EBAnO}}  

\begin{abstract}
Despite the high accuracy offered by state-of-the-art deep natural-language models (e.g. LSTM, BERT), their application in real-life settings is still widely limited, as they behave like a black-box to the end-user.
Hence, explainability is rapidly becoming a fundamental requirement of future-generation data-driven systems based on deep-learning approaches.
Several attempts to fulfill the existing gap between accuracy and interpretability have been done. 
However, robust and specialized xAI (Explainable Artificial Intelligence) solutions tailored to deep natural-language models are still missing.

\noindent
We propose a new framework, named \XXX, which provides innovative prediction-local and class-based model-global explanation strategies tailored to black-box deep natural-language models.
Given a deep NLP model and the textual input data, \XXX\ provides an objective, human-readable, domain-specific assessment of the reasons behind the automatic decision-making process.
Specifically, the framework extracts sets of \textit{interpretable features} mining the inner knowledge of the model. Then, it quantifies the influence of each feature during the prediction process by exploiting the novel \textit{normalized Perturbation Influence Relation} index at the local level and the novel \textit{Global Absolute Influence} and \textit{Global Relative Influence} indexes at the global level.
The effectiveness and the quality of the local and global explanations obtained with \XXX\ are proved on 
(i) a sentiment analysis task performed by a fine-tuned BERT model, and (ii) a toxic comment classification task performed by an LSTM model.

\end{abstract}


\end{frontmatter}

\definecolor{ansi-red}{HTML}{E75C58}
\definecolor{ansi-red-intense}{HTML}{B22B31}
\newcommand*\circled[1]{\textcircled{\raisebox{-0.9pt}{#1}}}
\newcommand{\XXX}{{\sc T-EBAnO}}

\section{Introduction}
\label{sec:intro}

Nowadays more and more deep-learning algorithms such as BERT~\cite{DBLP:journals/corr/abs-1810-04805} and LSTM~\cite{articleLSTM} are exploited as the ground basis to build new powerful automatic decision-making systems to solve complex natural language processing (NLP) tasks, e.g., text classification, question answering (QA), and sentiment analysis.
These models are often very accurate, even exceeding human performance (e.g. in \cite{wang2018glue, rajpurkar2016squad, NASEEM202058, BASIRI2021279}), but very opaque.
They are often defined black-boxes: given an input, they provide an output, without any human-understandable insight about their inner behavior.
Moreover, the huge amount of data required to train these black-box models is usually collected from people's daily lives (e.g. web searches, social networks, e-commerce) increasing the risk of inheriting human prejudices, racism, gender discrimination and other forms of bias \cite{Lepri2017, bolukbasi2016man}.
For these reasons, despite the promise of high accuracy, the applicability of these algorithms in our society and in real industrial settings is widely limited.
So, the demand for new Explainable Artificial Intelligence (xAI) solutions in future-generation systems is rapidly growing and xAI components will become, in the near future, a design requirement in most data-driven decision-making processes~\cite{deeks2019judicial}.

Table~\ref{tab:toxic-misleading-example} shows a clear example of a misleading prediction provided by an LSTM model\footnote{Details on the experiments leading to the reported result are provided in Section~\ref{ssec:exp-settings} trained to distinguish between \textit{Clean} and \textit{Toxic} comments.}.
\begin{table}
    \centering
    \small
    \begin{tabular}{|c|c|}
    \hline
    Sentence & $P(Toxic)$ \\ \hline
    \texttt{Politician-1 is an awesome man}   &  0.17 \\ \hline
    \texttt{Politician-1 is an intellectual}  &  0.89 \\ \hline
    \end{tabular}
    \caption{Misleading prediction example of a clean/toxic comment classification. The surname of a well-know politician is anonymized.}
    \label{tab:toxic-misleading-example}
\end{table}
In the example, both sentences are expressing \textit{Clean} language, however, the predictions are extremely contradictory and the black-box nature of the LSTM model does not allow us to understand why.
Thus, the complexity and the opacity of the learning process significantly reduces the adoption of those neural networks in real-life scenarios where a higher level of transparency is needed.
The new \textit{Explainable Artificial Intelligence} (xAI) field of research is currently trying to close the gap between model accuracy and model interpretability, to effectively increase the adoption of those models in real-life settings.

This work proposes \XXX\ (Text-Explaining BlAck-box mOdels), a novel explanation framework that allows understanding the decisions made by black-box neural networks in the context of \textit{Natural Language Processing}.\\
Human-readable prediction-local and model-global explanations are offered to users to understand why and how a prediction is made, hence allowing them to consciously trust the model's outcomes.


\XXX\ produces prediction-local explanations through a perturbation process applied on different sets of \textit{interpretable features}, i.e. parts of speech, sentences, and multi-layer word-embedding clusters, which are accurately selected to be meaningful for the model and understandable by humans. 
Then, \XXX\ evaluates the performance of the model in presence of the perturbed inputs, quantifying the contribution that each feature had in the prediction process through qualitative and objective indexes.
The proposed explanations enable end-users to decide whether a specific local prediction made by a black-box model is reliable and to evaluate the general behavior of the global model across predictions.
Prediction-local and model-global explanations are summarized in reports consisting of \textit{textual} and \textit{quantitative} contributions, allowing both expert and non-expert users to understand the reasons why a certain decision has been taken by the model under analysis.\\
As a case study, \XXX\ has been applied to explain 
(i) the well-known state-of-the-art language model BERT~\cite{DBLP:journals/corr/abs-1810-04805} applied in a sentiment analysis task, and 
(ii) a custom LSTM \cite{articleLSTM} model trained to solve a toxic comment binary classification task, i.e. detecting whether a document contains threats, obscenity, insults, or hate speech. 
Experimental results show the effectiveness of \XXX\ in providing human-readable, local vs global, interpretations of different model outcomes.\\
The novel contributions of the current work are provided in the following.
\begin{itemize}
    \item The design and development of a new xAI methodology, named \XXX, tailored to NLP tasks, to produce both prediction-local and model-global explanations, consisting of textual and numerical human-readable reports.
    \item The design of effective strategies to describe input textual documents through a set of model-wise interpretable features exploiting specific inner-model knowledge (Section~\ref{ssec:interpretable-feature-exctraction}). 
    \item The definition of quantitative and qualitative explanations, measuring the influence of each set of features on the local outcome provided by the black-box model (Section~\ref{ssec:local-explanation}).
    \item The definition of an innovative model-global explanation strategy, analyzing the influence of inter- and intra- class concepts, based on two new metrics, the \textit{Global Absolute Influence} and the \textit{Global Relative Influence} scores (Section~\ref{ssec:computing-global-insights}).
    \item A thorough experimental evaluation has been performed on two well-known black-box neural-network architectures, BERT and LSTM, on different textual data collections (Section~\ref{sec:experimental-results}).
\end{itemize}

The paper is organized as follows.
Section~\ref{sec:related-work} discusses xAI literature, 
Section~\ref{sec:architecture} provides an overview of the proposed solution, Section~\ref{sec:interpretable-features} provides the details about the \textit{interpretable features} compute by our framework, and Section~\ref{sec:explanations} describes how the local and global explanations are computed.
Section~\ref{sec:experimental-results} presents the experimental results and discusses the prediction-local and model-global explanation reports produced by \XXX. 
Finally, Section~\ref{sec:conclusion} concludes this work and presents future research directions.

\section{Literature review}
\label{sec:related-work}

Research activities in xAI can be classified based on \cite{Guidotti:2018:SME:3271482.3236009, samek2019, 8466590} 
data-type (e.g. structured data, images, texts), 
machine learning task (e.g. classification, forecasting, clusterization), 
and characteristics of the explanations (e.g. local vs global).
More generally, explanation frameworks can be grouped in (i) 
model-agnostic, 
(ii) domain-specific, and 
(iii) task-specific approaches.

Up to now, many efforts have been devoted to explain the prediction process in the context of structured data (e.g. measuring quantitative input influence~\cite{7546525}, by means of local rules in~\cite{Pastor:2019:EBB:3297280.3297328}) and of deep learning models for image classification (e.g. \cite{Selvaraju2019}, \cite{Fong_2017}, \cite{DBLP:conf/adbis/VenturaCG18}), while less attention has been devoted to domain-specific explanation frameworks for textual data analytics.

\vspace{0.2cm}
\noindent
\textbf{Model-agnostic approaches.} Tools like \cite{DBLP:journals/corr/RibeiroSG16} or \cite{NIPS2017_7062} can be applied to explain the decisions made by a black-box model on unstructured inputs (e.g. images or texts) and they provide interesting and human-readable results.
LIME~\cite{DBLP:journals/corr/RibeiroSG16} is a model-agnostic strategy that allows a local explanation to be generated for any predictive model. It approximates the prediction performed by the model with an interpretable model built locally to the data object to be predicted.
SHAP~\cite{NIPS2017_7062}, instead, is a unified framework able to interpret predictions produced by any machine learning model, exploiting a game-theoretic approach based on the concept of \textit{Shapley Values} \cite{shapley1953value}, 
by iteratively removing possible combinations of input features and measuring the impact that the removal of the features has over the outcome of the prediction task.
Since the above-mentioned techniques are \textit{model-agnostic}, they might not fully exploit the specific characteristics of the data domain and the latent semantic information specifically learned by the predictive models when computing an explanation. 
Although they can be applied in the context of NLP, they do not provide \textit{inner-model awareness}, i.e., they are not able to deeply explain what the model has specifically learned, since they do not exploit such information in their explanation process, leading to less specific explanations.
Moreover, in the specific case of NLP, model-agnostic techniques analyze the impact of singular words over the prediction, without taking into account the complex semantic relations that exist in textual documents (i.e., semantically correlated portions of text) and that is actually learned by modern neural networks. Also, perturbing singular words can have a very limited impact over the prediction process, in particular when dealing with long texts, other than being very computationally intensive, compromising the quality of the explanations.

\XXX\ addresses such limitations and is able to increase the precision of the produced explanations and to limit the feature search space by 
i) using \textit{domain-specific} feature extraction techniques and
ii) exploiting the \textit{inner knowledge} of the neural network to identify meaningful inter-word relations learned by the NLP model.

\vspace{0.2cm}
\noindent
\textbf{Domain-specific approaches.} An exhaustive overview of the existing xAI techniques for NLP models, applied in different contexts, such as social networks, medical, and cybersecurity, is presented in \cite{10.1007/978-3-030-22868-2_90}.
In the explanation process, many works exploit feature-perturbation strategies, analyzing the model reactions to produce prediction-local explanations, like in \cite{alvarez2017causal, DBLP:conf/adbis/VenturaCG18, NIPS2017_7062, DBLP:journals/corr/RibeiroSG16, li2016understanding, murdoch2017automatic}.
This straightforward idea is very powerful but requires a careful selection of the input features to be perturbed.\\
Differently from \textit{model-agnostic} and \textit{domain-agnostic} frameworks \cite{DBLP:journals/corr/RibeiroSG16, NIPS2017_7062}, some strategies have been explored by \textit{domain-specific} works to determine the information contained in the target model, with the aim to select the most relevant features to be perturbed. 
The feature extraction process is of utmost importance in the explanation process since the quality of the produced explanations strictly depends on this step.
In \cite{murdoch2017automatic} the authors
propose the use of an approximate brute-force strategy to analyze the impact that phrases in the input text have over the predictions made by LSTM models. Also, they define an importance score, that exploits the parameters learned by an LSTM model, to select the phrases which consistently provided a large contribution in the prediction of a specific class.  
However, this approach has been tailored to LSTM models, making it difficult to generalize the solution.
In \cite{alvarez2017causal} the authors proposed an explanation strategy tailored to structured and sequential data models with a perturbation strategy that exploits the training of a \textit{variational autoencoder} to perturb the input data with semantically related variations, introducing controlled perturbations. However, this explanation strategy has been mainly focused on explaining sequence to sequence scenarios (e.g. machine translation), and the perturbation requires the training, in advance, of a variational autoencoder model, introducing a further level of opacity and complexity in the explanation process.
The authors in \cite{lei2016rationalizing} propose to learn how to explain a predictive model jointly with the training of the predictor. 
To this aim, they introduce an \textit{encoder-generator} framework able to extract a subset of inputs from the original text as an interpretable summary of the prediction process. 
Again, the training of a separate model is required to extract the whole explanation, making also this solution equivocal for the end-user.
The authors in \cite{li2016understanding} proposed an explanation process based on a novel strategy to select the minimal set of words to perturb what causes a change in the decision of the model. To this aim, a reinforcement learning approach has been exploited.
However, as in previous cases, this method requires the training of an external model to extract features to be perturbed, increasing the complexity and affecting the reliability of the explanation process itself.

Differently than the above-mentioned works, \XXX\ implements a feature-extraction process that exploits the specific information learned by the predictive model, without the need to train external resources. 
\XXX\ exploits the embedding representation of the textual input data, available in the inner layer of the neural network, to identify correlated portions of input text accordingly to the model, which are used in the explanation process.
To support this choice, we recall that textual embeddings have interesting interpretable properties, as described in \cite{trifonov-etal-2018-learning}.
Following the insights discussed by the authors in \cite{Ethayarajh2019HowCA}, modern natural-language models incorporate most of the context-specific information in the latest and inmost layers. 
\XXX\ exploits the textual embedding representations as interpretable features to explain model outcomes. 

\vspace{0.2cm}
\noindent
\textbf{Task-specific approaches.} Finally, not every task can be explained with model-agnostic or domain-specific approaches. This is why interpretable task-specific solutions are also relevant.
In \cite{ZHOU2020} the authors focused their attention on explaining the duplicate question detection task developing a specific model based on the attention mechanism, proposing to interpret the model results by visually analyzing their attention matrix to understand the inter-words relations learned by the model. However, exploiting attention can be performed only for black-box models that are based on this mechanism, and it can be hard to interpret for non-expert users.
The authors of \cite{ZHENG2019170} developed an explainable tag-based recommendation model that increases the interpretability of its results by proposing an overview of user’s preference correlated with learned topics and predicted tags, but without actually focusing on the reliability of the model or on the possible presence of bias.
In \cite{LUGHOFER201716} the authors introduced a specific linguistic explanation approach for fuzzy classifier decisions, which are shown in textual form to users. They focus on a high abstraction level of explanations providing reasons, confidence, coverage, and feature importance. However, their approach does not take into account the complexity of deep learning models.
In \cite{KHODABANDEHLOO2020} the authors propose a framework for recognizing symptoms of cognitive decline that provides natural language explanations of the detected anomalies generated from a trained tree regression algorithm. However, this solution is customized for this specific task and not easily extendable to other contexts.

\XXX\ proposes a new local and global explanation process for state-of-the-art deep NLP models.
\XXX\ fills in the gap of missing customized solutions for explaining deep NLP models, by introducing a novel architecture and experimental section. 
Specifically we introduce 
(i) a novel feature extraction process specifically tailored to textual data and deep natural language models, 
(ii) new perturbation strategies, and
(iii) novel class-based global explanations.

\section{\XXX\ overview}
\label{sec:architecture}

\begin{figure*}[!ht]
    \centering
    \includegraphics[width=.8\textwidth]{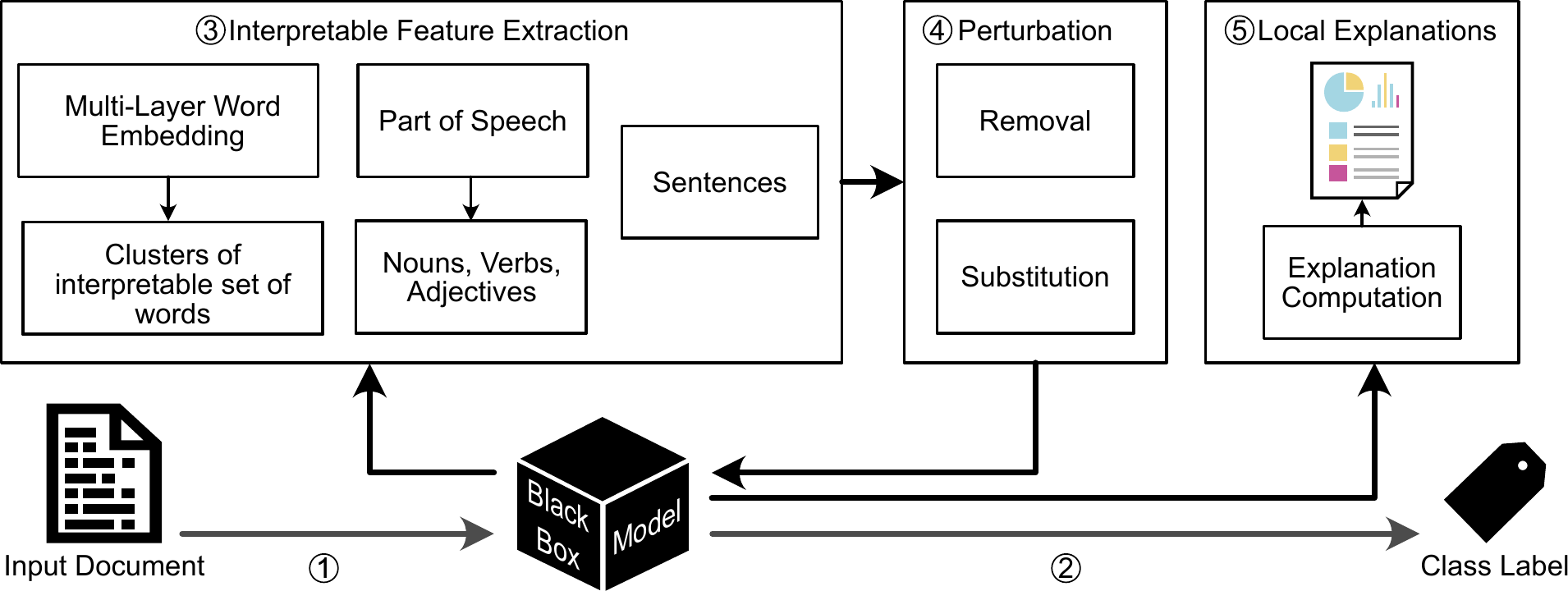}
    \caption{\XXX\ local explanation process.}
    \label{fig:t-ebano-arc}
\end{figure*}
\XXX\ explains the inner functionalities of black-box models in the context of NLP analytics tasks. Its architecture is shown in Figure~\ref{fig:t-ebano-arc} and includes different building blocks. 
Both \textit{model-agnostic} (i.e., part of speech, sentences) and \textit{model-aware} (i.e., multi-layer word embeddings) features are extracted by \XXX.

Given a classification tasks, an input document is provided to the black-box model \circled{1}, and the pre-trained model outputs its predicted \textit{class label} \circled{\small{2}}.
\XXX\ extracts a set of \textit{interpretable features} \circled{3} by exploiting either NLP techniques or the analysis of the knowledge hidden in the model itself (Section~\ref{ssec:interpretable-feature-exctraction}).
Then, it performs the \textit{perturbation} of the set of interpretable features and tests the outcomes of the model on the perturbed inputs \circled{4} (Section~\ref{ssec:perturbation}).
The perturbation of the interpretable features can influence the model outcome in different ways, as described in the following: 
\begin{itemize}
\item 
Case (a): the probability of the class under analysis \textit{increases}. It means that the analyzed features were negatively impacting on the process;
\item
Case (b): the predicted probability \textit{decreases}. It means that the perturbed features were positively impacting the class under analysis;
\item 
Case (c): the predicted probability \textit{remains roughly unchanged}. It means that the portion of input is irrelevant to the predictive model under analysis. 
\end{itemize}
The significance of the difference in the prediction process before and after the perturbation is evaluated through the $nPIR$ index, a quantitative metric to estimate the effect of the perturbation strategy (Section~\ref{ssec:quantitative-exp}).
Thus, \XXX\ generates the \textit{local explanation} report \circled{\small{5}}, 
showing the results of the analysis of the perturbations through an informative dashboard.

Finally, aggregating the local explanations produced for a corpus of input documents, \XXX\ provides model-global explanations highlighting relevant inter- and intra- class semantic concepts that are influencing the black-box decision-making process at a model-global level (details are provided in Section~\ref{ssec:computing-global-insights}).

\section{Interpretable features}
\label{sec:interpretable-features}

This Section describes the interpretable feature extraction (Section \ref{ssec:interpretable-feature-exctraction}), 
with a specific focus on the Multi-layer Word Embedding technique (Section \ref{sssec:mlwe}), 
and the feature perturbation (Section \ref{ssec:perturbation}) approaches adopted by \XXX.

\subsection{Interpretable feature extraction}
\label{ssec:interpretable-feature-exctraction}

The interpretable feature extraction block identifies meaningful and correlated sets of words (tokens) having an influence on the outcomes of the NLP model under the exam. 
It represents the most critical and complex phase in the explanation process workflow.
A set of words is meaningful for the model if its perturbation in the input document produces a meaningful change in the prediction outcome. 
On the other hand, a set of words is meaningful for a user if s/he can easily understand and use it to support the decision-making process.

\XXX\ considers both word (tokens) and sentence granularity levels to extract the set of interpretable features. 
Moreover, \XXX\ records the position of the extracted features in the input text, since the context in which words appear is often very important for NLP models.

\XXX\ includes three different kinds of \textit{interpretable feature extraction} techniques:

\begin{enumerate}
    \item 
    \textit{Multi-layer Word Embedding} (MLWE) feature extraction. 
    This strategy is the most powerful technique since it exploits the inner knowledge learned by the model to perform the prediction. 
    To access the inner knowledge of the network, this technique needs to know the inner details of the model under analysis. However, the process can be easily adapted to be compliant with different deep architectures (e.g., as reported in \cite{DBLP:conf/adbis/VenturaCG18}) and their hidden layers. 
    A detailed description of the MLWE feature extraction technique is provided in Section~\ref{sssec:mlwe}.
    
    \item
    \textit{Part-of-Speech} (PoS) feature extraction. 
    This strategy explores the semantic meaning of words by looking to which part-of-speech they belong to (e.g. nouns, adjectives).
    The intuition behind this type of feature extraction is that the semantic difference corresponding to distinct parts-of-speech can differently influence the model outcome. 
    Firstly, the input text is tokenized, leading to three features: the \textit{token} itself, its \textit{position} in the text, and its \textit{pos-tag} (i.e. part-of-speech tagging).
    Then, tokens are divided into correlated groups: adjectives, nouns, verbs, adverbs and others.
    Each group is considered as a separate \textit{interpretable feature} by \XXX\ in the perturbation phase.
    
    \item 
    \textit{Sentence-based} feature extraction. 
    This strategy considers each sentence separately to assess their influence on the model decisions.
    The straightforward intuition behind this strategy is to verify if the model captures the complete meaning of a sentence and uses it to derive the outcome.
    The \textit{sentence} feature extraction characterizes the input text with the \textit{position} of the sentence and the \textit{sentence} itself.
\end{enumerate}

Then, separately for each feature extraction method, \XXX\ tests \textit{pairwise combinations of features} to create larger groups of tokens corresponding to more complex concepts.
This allows us to efficiently explore a wider search space of interpretable features allowing us to find even more relevant prediction-local explanations.

\subsection{Multi-layer Word Embedding (MLWE) feature extraction}
\label{sssec:mlwe}

\begin{figure*}[!ht]
    \centering
    \includegraphics[width=.8\textwidth]{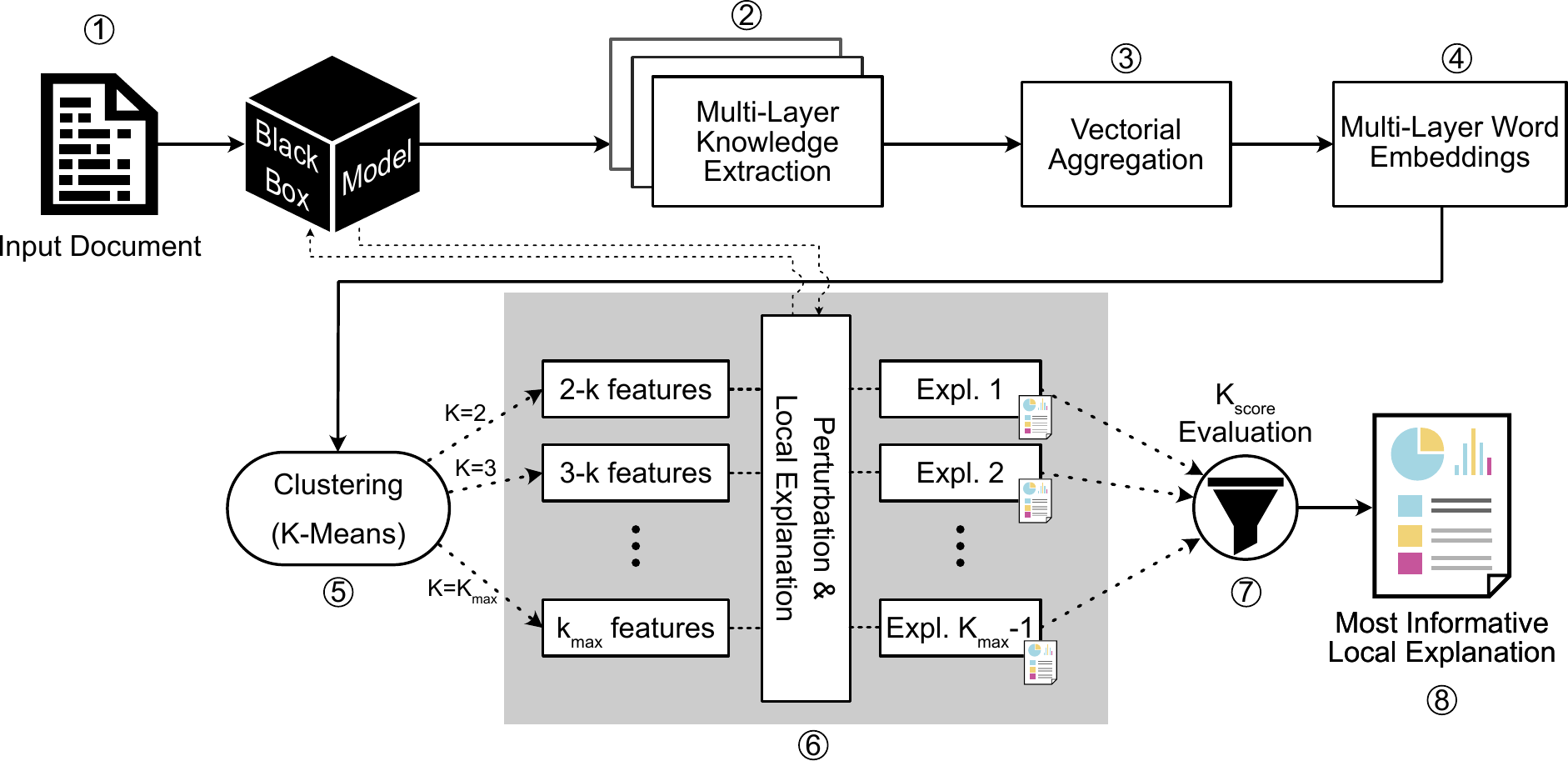}
    \caption{\XXX\ MLWE feature extraction process.}
    \label{fig:MLWE-process}
\end{figure*}


Deep Neural networks are trained to extract knowledge from training data learning a complex numerical model spreading this knowledge on multiple hidden layers.
During the prediction process of previously unseen data, all these layers contribute to the outcome.
Thus, to get a reliable explanation, it is necessary to mine all the knowledge hidden along with the layers of the model.
Thanks to the Multi-layer Word Embedding (MLWE) feature extraction, \XXX\ can achieve this goal.

First, \XXX\ analyzes the outcomes of multiple hidden layers to extract the numerical representation of the input at different levels of the network.
The \textit{Multi-layer Word Embedding} feature extraction process is shown in Figure~\ref{fig:MLWE-process}. 
Firstly, given an input document \circled{1}, a tensor containing the numerical embedding representations of different words in different layers is extracted \circled{2}.
Then, the intermediate embeddings of each layer are aggregated (e.g. through average or sum) and their dimensionalities are further reduced through PCA to obtain an embedded vectorial representation for each input word \circled{3}.
The outcomes of the aggregation step is the \textit{Multi-layer Word Embedding} representation of the input document \circled{4}.
The intuition is that words with a similar MLWE are considered highly correlated by the model and, if grouped together, they represent key input concepts, that most probably are influencing the current prediction.
The MLWE feature extraction, and in particular the extraction of the aggregated word embeddings from multiple layers, has to be achieved in different ways depending on the neural network architecture under the exam.
Further details about MLWE feature extraction tailored to LSTM and to BERT are provided in Sections~\ref{sssec:exp-sec-lstm-mlwe} and \ref{sssec:exp-sec-bert-mlwe} respectively.



Once the MLWEs are extracted, they are analyzed through an unsupervised clustering analysis \circled{5} to identify sets of correlated words that share common behaviors inside the model under exam. 
The aim of the unsupervised analysis it to identify the smallest groups of input words that have the highest impact over the model outcome.
This allows also to reduce the search space, without affecting the quality of the features.
For this purpose, \XXX\ exploits the K-Means~\cite{1056489} clustering algorithm, since it provided good performance in a similar context \cite{DBLP:conf/adbis/VenturaCG18} and represents a good trade-off with computational time.
A critical parameter when dealing with K-Means is the setting of the desired number of groups $K$ to correctly model interesting subsets of data.
\XXX\ applies K-Means to identify a number of groups ranging in $[2,K_{max}]$, where the max number of clusters $K_{max}$ is a function of the input size and has been empirically set to:
\begin{equation}
    K_{max} = \sqrt{N_{words} + 1}
\end{equation}
On the one hand, using small fixed values of $K$ with large input texts leads to large clusters of words containing both influential and less impacting words, and consequently the explanation provided will be of low interest.
On the other hand, the number of words $N_{words}$ in a text can be very high and it would not be feasible neither useful to evaluate partitioning that take into account values of $K$ as large as the number of words $N_{words}$.
For this reason, evaluating a number of clusters $K$ that is at most equal to the root of the number of words $N_{words}$ in a text allows to maintain a good trade-off between partitioning size and performance.
\XXX\ produces a \textit{quantitative explanation} for each $K$ \circled{6}, as detailed in Section~\ref{ssec:local-explanation}.
For each value of $K \in [2,K_{max}]$, $K$ perturbations will be analyzed.
In this way, a large number of potentially useful \textit{local explanations} are produced by \XXX.

The objective, however, is to provide only the best explanation to the end-user. \XXX\ selects the \textit{most informative local explanations} as those extracting the most knowledge from the behavior of the model over a single prediction. 
To this aim, for each value of $K$, a score is computed \circled{7} by means of the \textit{normalized Perturbation Influence Relation} (nPIR) index (introduced in Section~\ref{ssec:local-explanation}), which is computed as follows:
\begin{equation}
\label{eq:k_score}
    K_{score} = \max\limits_{\kappa \in K} \left(\frac{nPIR_{\kappa }}{|\kappa|}\right) - \min\limits_{\kappa \in K} \left(\frac{nPIR_{\kappa }}{|\kappa|}\right)
\end{equation}
Where \textit{$\kappa$} is the current cluster, $|\kappa|$ is the number of words inside the cluster and $nPIR_{\kappa}$ is the $nPIR$ value of the current cluster \textit{$\kappa$}, that measures the positive or negative influence of perturbing the tokens in $\kappa$ (further details are provided in Section~\ref{ssec:quantitative-exp}).
The selected set of features is the set with $\max(K_{score})$. 
The $K_{score}$ tends to assign a high influence to small clusters.
The range of $K_{score}$ is [0,2], where 2 is the most informative local explanation, obtained when MLWE finds a cluster $C_1$ composed by exactly one word with $nPIR_{\kappa=C_1}=1$ and another cluster $C_2$ composed exactly by one word with $nPIR_{\kappa=C_2}=-1$.
Instead, the least informative local explanation has a score of 0, and it is identified when, for instance, MLWE finds $K$ clusters of words all being neutral for the prediction of the class label, hence having $nPIR_{\kappa}=0$.

The output of this process is the \textit{most informative local-explanation} \circled{8}.
\subsection{Interpretable feature perturbation}
\label{ssec:perturbation}

After the extraction of the interpretable feature sets, a perturbation phase is performed by introducing noise and consequently assessing the impact of the perturbed features on the model outcomes.
Adding noise to the model input is a well-known technique adopted by different state-of-the-art approaches \cite{alvarez2017causal, DBLP:conf/adbis/VenturaCG18, NIPS2017_7062, DBLP:journals/corr/RibeiroSG16} to study the model behavior through the effects on the outcomes.
Different input data types require different perturbation strategies.
In case of textual data, the perturbation can be performed by \textit{feature removal} or \textit{feature substitution}. 

In the \textit{feature removal} approach provided by \XXX, all the interpretable features are iteratively removed from the input text, producing new perturbed variations of the input itself.
The perturbed variations of the input are then fed back into the model under analysis and its predictions are collected and analyzed by \XXX\ to produce the \textit{local explanation} report (see Section \ref{ssec:local-explanation}).
Examples of explanations produced by feature removal perturbation are shown in Figures~\ref{fig:toy-ex-adjective}, \ref{fig:toy-ex-sentence}, and \ref{fig:toy-ex-mlwe}.
From the input text in Figure~\ref{fig:toy-ex-original}, the words highlighted in Figure~\ref{fig:toy-ex-adjective}, the sentence highlighted in Figure~\ref{fig:toy-ex-sentence} and the words identified by MLWE in Figure~\ref{fig:toy-ex-mlwe} are removed.\\

The \textit{feature substitution} is also adopted by \XXX.
While the removal perturbation causes an absence of the concept associated with the removed words, the substitution perturbation introduces a new, possibly related, concept that can cause a change in the prediction. 
The \textit{feature substitution} perturbation requires an additional step to select new words that will replace the current ones. 
In \XXX, the substitution of words with their \textit{antonyms} is exploited.
This strategy turned out to be very powerful in some specific cases (e.g. Adjective-POS perturbation), but in general, it has several limitations: 
(i) some words can have many antonyms and the optimal choice might depend on the context, 
(ii) antonyms do not exist for some words (e.g. nouns), and
(iii) the choice of the new words to be inserted in the substitution of the feature is task-specific (e.g. antonyms work with opposite class labels like \textit{Positive} and \textit{Negative} in sentiment analysis, but are not suited with independent class labels as in topic detection).
Thus, the effectiveness of this perturbation strategy is affected by these limitations.
Figures~\ref{fig:toy-ex-adjective-removal} and \ref{fig:toy-ex-verb-removal} show two examples of explanations performed using this technique. 
For the Adjective-POS features, it is straightforward to find meaningful antonyms.
On the contrary, for Verb-POS features, the result is very difficult to evaluate, since verbs like \{{\fontfamily{cmtt}\selectfont was, have}\} are substituted with \{{\fontfamily{cmtt}\selectfont differ, lack}\}.
This feature perturbation strategy remains an open task left for further inspection in future works.

\section{Explanations}
\label{sec:explanations}

This Section presents the prediction-local (Section \ref{ssec:local-explanation}) and the model-global (Section \ref{ssec:computing-global-insights}) explanation processes implemented in \XXX.

\subsection{Prediction-local explanations}
\label{ssec:local-explanation}

\begin{figure}[t!]
    \centering
    \subfloat[Original text]{
        \label{fig:toy-ex-original}
        \fbox{\begin{minipage}{1\columnwidth}
        \scriptsize
        \fontfamily{cmtt}\selectfont
        This film was very awful. I have never seen such a bad movie.
        \end{minipage}}
    }\\
    \subfloat[\texttt{EXP1:} Adjective - POS feature extraction with removal perturbation.]{
        \label{fig:toy-ex-adjective}
        \fbox{\begin{minipage}{1\columnwidth}
        \scriptsize
        \fontfamily{cmtt}\selectfont
        This film was very \textcolor{ansi-red-intense}{awful}. I have never seen such a \textcolor{ansi-red-intense}{bad} movie.
        \end{minipage}}
    }\\
    \subfloat[\texttt{EXP2:} Sentence feature extraction with removal perturbation.]{
        \label{fig:toy-ex-sentence}
        \fbox{\begin{minipage}{1\columnwidth}
        \scriptsize
        \fontfamily{cmtt}\selectfont
        \textcolor{ansi-red-intense}{This film was very awful.}
            I have never seen such a bad movie.
        \end{minipage}}
    }\\
    \subfloat[\texttt{EXP3:} Multi-layer word embedding feature extraction with removal perturbation.]{
        \label{fig:toy-ex-mlwe}
        \fbox{\begin{minipage}{1\columnwidth}
        \scriptsize
        \fontfamily{cmtt}\selectfont
        This film \textcolor{ansi-red-intense}{was} very \textcolor{ansi-red-intense}{awful}. I have never seen such a \textcolor{ansi-red-intense}{bad} \textcolor{ansi-red-intense}{movie}
        \end{minipage}}
    }\\
    \subfloat[\texttt{EXP4:} Adjective-POS feature extraction with substitution perturbation.]{ 
        \label{fig:toy-ex-adjective-removal}
        \fbox{\begin{minipage}{1\columnwidth}
        \scriptsize
        \fontfamily{cmtt}\selectfont
        This film was very \textcolor{ansi-red-intense}{[awful] nice}. I have never seen such a \textcolor{ansi-red-intense}{[bad] good} movie
        \end{minipage}}
    }\\
    \subfloat[ \texttt{EXP5:} Verb-POS feature extraction with substitution perturbation.]{
        \label{fig:toy-ex-verb-removal}
        \fbox{\begin{minipage}{1\columnwidth}
        \scriptsize
        \fontfamily{cmtt}\selectfont
        This film \textcolor{ansi-red-intense}{[was] differ} very awful. I \textcolor{ansi-red-intense}{[have] lack} never seen such a bad movie.
        \end{minipage}}
    }
    
    \caption{Examples of a \textit{textual explanation} report. The original text was labeled by BERT as \textit{Negative} with a probability of 0.99. The most relevant features are highlighted in red. Removed features are in squared brackets.}
    \label{fig:toy-ex-textual-explanation}
    \qquad
    \begin{minipage}{\columnwidth}
        \centering
        \scriptsize
        \begin{tabular}{|c|c|c|c|c|}
            \hline
            Explanation & Feature $f$ & $L_o$ & $L_f$  & $nPIR_f(N)$ \\ \hline
             EXP1 & POS-Adjective & N & P & 0.998 \\ \hline
             EXP2 & Sentence & N & N & 0.000 \\ \hline
             EXP3 & MLWE & N & P & 0.984 \\ \hline
             EXP4 & POS-Adjective (sub.) & N & P & 0.999 \\ \hline
             EXP5 & POS-Verb (sub.) & N & N & 0.000 \\ \hline
        \end{tabular}
        \captionof{table}{Quantitative explanation for example in Figure~\ref{fig:toy-ex-textual-explanation}. P is the positive label, N is the negative label. The (sub.) suffix indicates that the substitution perturbation has been applied.}
        \label{tab:toy-ex-numerical-explanation}
\end{minipage}
    
\end{figure}

To produce the local explanations, \XXX\ exploits the outcomes of the model when fed with the original input and its perturbed versions.
A local explanation consists of two main parts: 
a \textit{textual explanation} (Figure~\ref{fig:toy-ex-textual-explanation}) and 
a \textit{quantitative explanation} (Table~\ref{tab:toy-ex-numerical-explanation}), as detailed in the following.

\vspace{0.1cm}
\noindent
\textbf{Textual explanation.} 
The \textit{textual explanation} highlights the most relevant sets of features for the model under analysis also allowing to understand the context in which they appear.
Many sets of features can be extracted for each interpretable feature extraction technique.
Figure~\ref{fig:toy-ex-textual-explanation} shows a simple example of textual explanations. 
For this example the BERT model has been trained to detect sentiment the sentiment of a textual document, either positive (P) or negative (N).
Given the input document in Figure~\ref{fig:toy-ex-original}, the model outputs a negative sentiment. So, the user can inspect the highlighted features (in red) in the textual explanations in Figures~\ref{fig:toy-ex-adjective}, \ref{fig:toy-ex-sentence}, \ref{fig:toy-ex-mlwe}, \ref{fig:toy-ex-adjective-removal}, and \ref{fig:toy-ex-verb-removal} to find out which are the most important sections of the input that have been exploited by the model to make its decision.

\vspace{0.1cm}
\noindent
\textbf{Quantitative explanation.}
\label{ssec:quantitative-exp}
The \textit{quantitative explanation} shows the influence of each set of extracted features separately for each prediction by evaluating the newely introduced \textit{nPIR} index (normalized Perturbation Influence Relation).
It assesses the importance of an input feature for a given prediction, analyzing its performance before and after the perturbation of a feature (or set of features) extracted from the input data.
The $nPIR$ for a class of interest is computed as follows:
\begin{gather}  
nPIR_f(c) =softsign(P_f(c)*b-P_o(c)*a)\\
a=1-\frac{P_o(c)}{P_f(c)}  ;  b=1-\frac{P_f(c)}{P_o(c)}
\end{gather}
where $c$ is the class of interest, 
$P_o$ is the probability of belonging to the class of interest analyzing the original input $o$,
and $P_f$ is the same probability analyzing the input with the perturbed feature $f$.
The coefficient $a$ is the contribution of input $o$ w.r.t. the perturbed input. 
Similarly, $b$ represents the contribution of the perturbation of $f$ w.r.t. the original feature.
The $PIR$ value is then normalized exploiting the well-known $softsign$ function in order to have a linear normalization of $PIR$ values close to 0 while smoothly converging to 1 in case of positive impact and to -1 in case of negative impact.
The $nPIR$ index values are in the range $[-1,1]$.
The higher the $nPIR_f$ (close to 1), the more the feature $f$ is positively influencing the class of interest.
On the opposite, the lower the $nPIR_f$ (close to -1), the more the feature $f$ is negatively influencing the class of interest. 
Exploiting the $nPIR$ index allows us to define a threshold to identify \textit{informative} explanations.
For instance, 
if $-0.5<nPIR_f<0.5$, then the difference between the probabilities before and after the perturbation of $f$ is not informative for a threshold of 0.5. 
Values of $nPIR_f<-0.5$ (or $nPIR_f>-0.5$) means that the perturbation of feature $f$ is contributing negatively (or positively) to the prediction, by decreasing (or increasing) significantly the probability of belonging to the class $c$.\\
The \textit{quantitative explanation}, is computed by \XXX\ for each feature extraction technique, for each set of features perturbation, and for each class that can be predicted by the model.


Table~\ref{tab:toy-ex-numerical-explanation} shows the \textit{quantitative explanations} for the \textit{textual explanations} in Figure~\ref{fig:toy-ex-textual-explanation}.
For each \textit{interpretable feature} $f$ the labels assigned by the model before and after their perturbation are reported in columns $L_o$ and $L_f$ respectively along with the $nPIR$ value calculated for the class-of-interest negative (N).
Perturbing the POS adjectives in Figure~\ref{fig:toy-ex-adjective} (EXP1) or the MLWE cluster in Figure~\ref{fig:toy-ex-mlwe} (EXP3) the $nPIR$ is very close to 1.
This means that these sets of features are very relevant for the model outcome: \textit{removing} one of these features will cause completely different outcomes from the model, changing the prediction from negative (N) to positive (P).
Instead, the perturbation of the sentence in Figure~\ref{fig:toy-ex-sentence} (EXP2) is not relevant at all for the model, showing a value of $nPIR$ equal to 0.
We can conclude that the feature sets \{{\fontfamily{cmtt}\selectfont awful, bad}\} and \{ {\fontfamily{cmtt}\selectfont was, awful, bad, movie} \} are the real reason why the model is predicting the negative class.
The information contained in the sentence \{ {\fontfamily{cmtt}\selectfont This film was very awful} \} instead does not justify the model outcome alone, like the rest of the text that is also contributing to the prediction.\\
The \textit{quantitative explanations} obtained through the substitution perturbation (EXP4 and EXP5) have been also reported in Table~\ref{tab:toy-ex-numerical-explanation}. 
Even from these results, it is evident that the substitution perturbation has great potential in expressiveness when it is possible to find suitable antonyms.
In the case of Adjective-POS substitution (EXP4), the quantitative explanation shows a $nPIR$ value close to 1.
On the contrary, in the case of EXP5, verbs are replaced with semantically incorrect words (not antonyms) in the context of the phrases, showing no impact in the prediction process with a $nPIR$ equal to 0.

\subsection{Per class model-global explanation}
\label{ssec:computing-global-insights}

\XXX\ is able to provide per-class \textit{model-global} explanations of the prediction process.
The local explanations computed for a corpus of input documents are aggregated and analyzed together, highlighting possible misleading behaviors of the predictive model.

\begin{figure}[t!]
    \centering
    \subfloat[Absolute influential words.]{
    \label{fig:absolute-words}
        \includegraphics[width=0.22\textwidth]{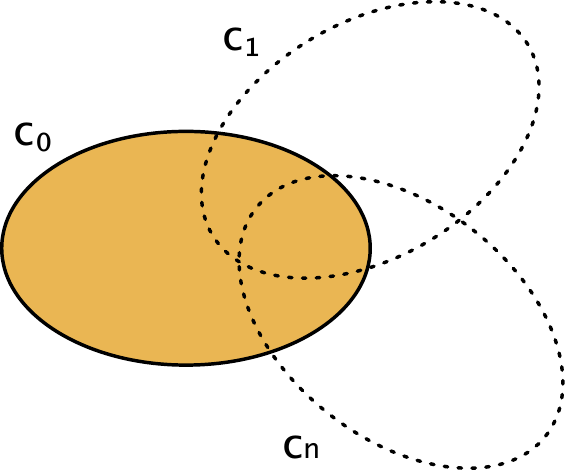}
    }
    \subfloat[Relative influential words.]{
    \label{fig:relative-words}
        \includegraphics[width=0.22\textwidth]{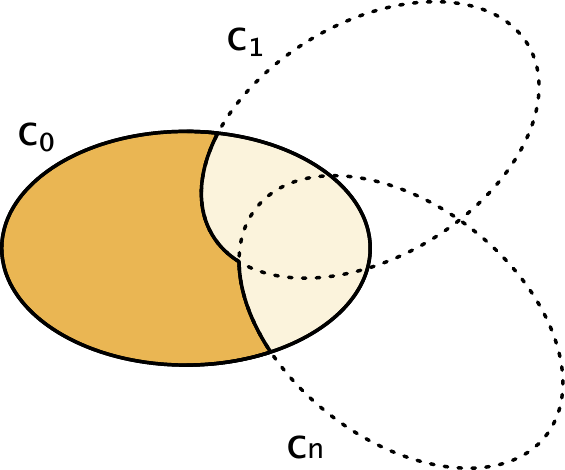}
    }
    \caption{Influential set of words at model-global level for class-of-interest $c_0$.}
    \label{fig:absolute-relative-words}
\end{figure}

Two indices have been introduced to measure the global influence of the corpus of input documents: 
(i) the \textit{Global Absolute Influence} ($GAI$) described by  Algorithm~\ref{alg:GlobalAbsoluteScoreAlg}, and 
(ii) the \textit{Global Relative Influence} ($GRI$) defined in Equation~\ref{eq:GlobalRelativeScoreEq}.
The $GAI$ score measures the global importance of \textit{all} the words impacting on the class-of-interest, without distinction concerning other classes (Figure~\ref{fig:absolute-words}). 
On the other hand, the $GRI$ score evaluates the relevance of the words influential \textit{only} (or mostly) for the class-of-interest, differently from other classes (Figure~\ref{fig:relative-words}).

Analyzing $GAI$ and $GRI$ scores, the user can extrapolate which are the most relevant inter- and intra- class semantic concepts that are affecting the decision-making process at a model-global level. 
For example, if a word is influential for all the possible classes, it will have a high $GAI$ score and a $GRI$ score close to 0.
On the contrary, a word might show a high $GRI$ score for a specific class, while having a very low $GAI$, meaning that it is influential and differentiating for that specific class only.

The global explanations are computed for each available class $c \in C$, analyzing the set of local explanations $E$.
The local explanations $e_{d,f} \in E$ are computed for each document $d \in D$ and for each interpretable feature $f$.

\vspace{0.2cm}
\noindent
\textbf{Global Absolute Influence.} The Global Absolute Influence value is computed following the process described in Algorithm \ref{alg:GlobalAbsoluteScoreAlg}. Given a set of local explanations $E$ generated for a corpus of documents $D$, the algorithm computes the global score for each possible class-of-interest and for each lemma (base form of a word) contained in the most informative local explanations.
Only MLWE explanations are exploited in the algorithm (line~\ref{alg:ln:getExplanationsMLWE}) since it is the only feature extraction strategy that exploits inner model knowledge (see Section~\ref{sssec:mlwe}).
Then, given the MLWE explanations related to a document $e_{d}$, only the most influential one $\hat{e}_{d}$, i.e., the one with the highest $nPIR$, is selected (line~\ref{alg:ln:getMostInfluentialExplanation}) and the lemmas $L_{\hat{e}_{d}}$ are extracted from the tokens contained in the corresponding interpretable feature (line~\ref{alg:ln:lemmatizeTokens}).
The algorithm analyzes \textit{lemmas} instead of \textit{tokens} (words) in order to group together their inflected forms, obtaining more significant results.
Finally, the GAI score for the corresponding class-of-interest $c$ and lemma $l$ is updated (line~\ref{alg:ln:GAI}) by summing the $nPIR$ score of the the explanation $\hat{e}_{d}$, only if it is positively impacting the prediction (i.e., if $nPIR>0$).
The output of the algorithm is the set of \textit{Global Absolute Influence} scores. 

The $GAI$ score will be 0 for all the lemmas that have always brought a negative influence on class $c$, and it will grow proportionally to the frequency and to the positive influence of each lemma positively influencing class $c$. The higher the GAI score, the most positively influential a lemma is for the model under analysis with respect to class $c$.

\vspace{0.2cm}
\noindent
\textbf{Global Relative Influence.} The Global Relative Influence score highlights the most influential and differentiating lemmas for each class-of-interest, discarding lemmas with multiple impact on other classes.
The $GRI$ for a class-of-interest $c$ and for a specific lemma $l$ is defined as:
\begin{equation}
\label{eq:GlobalRelativeScoreEq}
    GRI(c,l) = Max[0, GAI(c,l) - \sum_{c_i \neq c}^{C} GAI(c_i,l)]
\end{equation}

The $GRI$ score is 0 when a lemma is more relevant for other classes than for the one under exam, while $GRI>0$ if its influence is higher for class $c$ than all the other classes. 
The higher the GRI value, the more specific the lemma influence is with respect to the class-of-interest.

Section~\ref{ssec:global-insights} provides an experimental analysis of the insights provided by \XXX\ at a model-global level.

\begin{algorithm}[ht!]
\SetKwInput{KwInput}{Input}           
\SetKwInput{KwOutput}{Output}
\small
\SetAlgoLined
\LinesNumbered
\KwInput{Local explanations $E$, Classes $C$ . }
\KwOutput{\textit{$GAI$} scores for each class label and lemma.}
$GAI \leftarrow$ initHashMap($0$)\;
$E_{\text{MLWE}}\leftarrow$ getExplanationsMLWE($E$)\; \label{alg:ln:getExplanationsMLWE}
\For{$c$ \textbf{in} $C$}  {
    \For{$e_{d}$ \textbf{in} $\{E_{\text{MLWE}} : 0 \le d \le |D|\}$}  {
        $\hat{e}_{d} \leftarrow$ getMax\_nPIR\_Explanation($e_{d}$,$c$)\; \label{alg:ln:getMostInfluentialExplanation}
        $L_{\hat{e}_{d}} \leftarrow$ lemmatizeTokens($\hat{e}_{d}.featureTokens$)\; \label{alg:ln:lemmatizeTokens}
            \For{$l$ \textbf{in} $L_{\hat{e}_{d}}$} {
                     $GAI(c,l) \leftarrow GAI(c,l) + Max[0, \hat{e}_{d}.nPIR$]\; \label{alg:ln:GAI}
            }
     }
 }
\textbf{return} $GAI$\;
 \caption{Global Absolute Influence.}
 \label{alg:GlobalAbsoluteScoreAlg}
\end{algorithm}

\section{Experimental results}
\label{sec:experimental-results}
In this Section we present the experiments performed to assess the ability of \XXX\ to provide useful and human-readable insights on the decisions made by a black-box NLP model.

\subsection{Use cases}
\label{ssec:exp-settings}

We applied \XXX\ in two different binary text-classification use cases, as described in the following, intending to address diverse state-of-the-art NLP models, specifically LSTM and BERT, to show the flexibility of \XXX\ independently of the specific black-box model.

\vspace{0.2cm}
\noindent
\textbf{Use case 1.} The first task is a binary \textit{toxic comment classification}, and it consists of predicting whether the input comment is \textit{clean} or \textit{toxic}, i.e., it contains inappropriate content such as obscenity, threat, insult, identity attack, and explicit sexual content.
An LSTM model applied to a civil comments dataset~\cite{DBLP:journals/corr/abs-1903-04561} has been used. 
The \textit{toxic} class label contains several subtypes of toxic comments as identity attacks, insults, explicit sexuality, and threats.  
The LSTM model is composed by an embedding 300-dimensional layer, two bidirectional LSTM layers (with 256 units for each direction), and finally, a dense layer with 128 hidden units. 
Transfer learning has been exploited using GloVe \cite{pennington2014glove} (with 300-dimensional vectors) for the embedding layer.
After training, the custom LSTM model reached an accuracy of 90\%.

\vspace{0.2cm}
\noindent
\textbf{Use case 2.} The second selected task is \textit{sentiment analysis}, and it consists of predicting if the underlying sentiment of an input text is either positive or negative. The BERT base (uncased) pre-trained model \cite{DBLP:journals/corr/abs-1810-04805} has been chosen as the black-box predictive model, and it has been applied to the \textit{IMDB dataset} \cite{maas-EtAl:2011:ACL-HLT2011}, which is a reference set of data for sentiment analysis.
We performed a fine-tuning step of the BERT model~\cite{DBLP:journals/corr/abs-1810-04805} by adding a classification layer on top of the last encoder transformer's stack.
The BERT model, fine-tuned on the IMDB textual reviews, reached an accuracy of 86\%.




\subsection{Multi-layer word embedding}
\label{sssec:exp-sec-lstm-mlwe}
\label{sssec:exp-sec-bert-mlwe}

\vspace{0.2cm}
\noindent
\textbf{LSTM.}
RNN with LSTM units are robust architectures that can learn both the time sequence dimension and the feature vector dimension. 
Multiple LSTM layers usually characterize them, and they can take as input an embedded representation of the text. 
As highlighted in Section~\ref{ssec:exp-settings}, the developed LSTM model exploits one embedding layer that works with full tokens and two bidirectional LSTM layers.
For these reasons, the MLWE exploits the single embedding layer to extract a tensor of shape $(t \times 300 \times 1)$. 
Then, a Principal Component Analysis is used to reduce the embedding matrix shape to $(t \times c)$ obtaining the \textit{multi-layer word embedding} representation for the custom LSTM model.


\begin{figure*}[ht!]
    \centering
    \includegraphics[width=.9\textwidth]{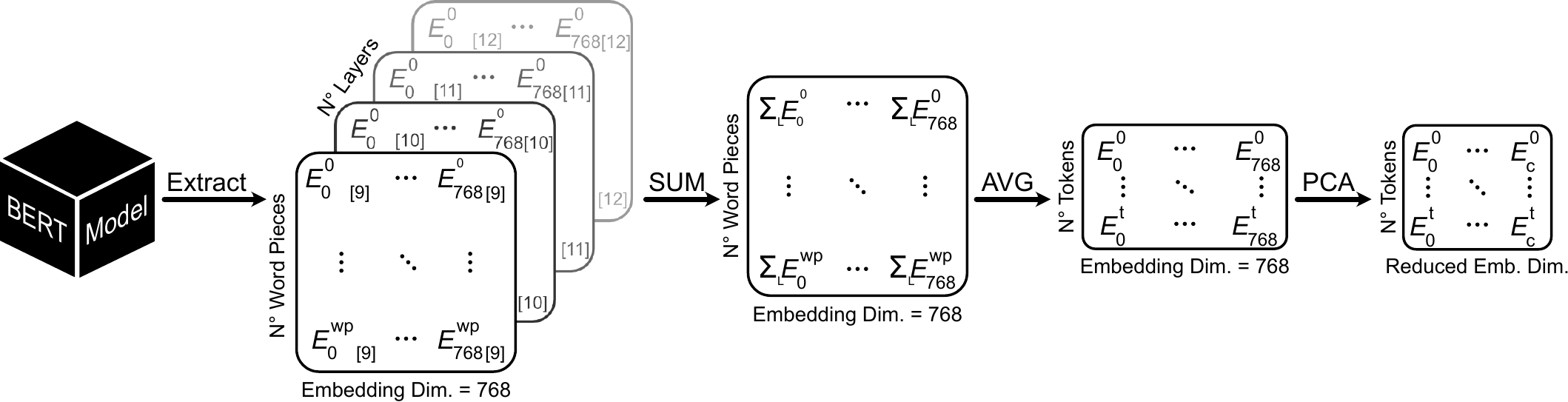}
    \caption{BERT multi-layer word embedding feature extraction process. With: $E_{c \text{ } [L_{id}]}^{<wp \text{ or } t>}$ such that: $E$ is the word embedding matrix, $wp$ and $t$ indicate the position of the word-piece and token respectively in the input text, $c$ is the component of the word embedding vector and $L_{id}$ is the layer from which is extracted. }
    \label{fig:mlwe_feature_extraction}
\end{figure*}

\vspace{0.2cm}
\noindent
\textbf{BERT.}
Figure~\ref{fig:mlwe_feature_extraction} shows all the steps of the \textit{multi-layer word embedding} (MLWE) feature extraction process in BERT. 
The \textit{base} version of the BERT model \cite{DBLP:journals/corr/abs-1810-04805} is composed by 12 transformer layers \cite{DBLP:journals/corr/VaswaniSPUJGKP17}, each producing an output of shape $(wp \times 768)$, where $wp$ is the number of word pieces extracted by BERT in its pre-processing phase.
The MLWE, in this case, analyzes the word embeddings extracted from the last four transformer layers of the model. It has been motivated in literature \cite{Ethayarajh2019HowCA} that modern natural language models incorporate most of the context-specific information in the last and deepest layers. 
Thus, the joint analysis of these layers allows the MLWE to extract features more related to the task under exam, avoiding too specific (if analyzing only the last layer) or too general (if analyzing only the firsts layers) word embeddings.

In the first step of the MLWE feature extraction, the last four transformer layer outputs (i.e. $L_{9}$, $L_{10}$, $L_{11}$, $L_{12}$) are extracted (Figure~\ref{fig:mlwe_feature_extraction}-left), resulting in a tensor of shape $(wp \times 768 \times 4)$.
Each row is the embedding representation for each word piece in each layer.
Then, the outputs of the four layers are aggregated summing the values of the embeddings over the layer axes in a matrix of shape $wp \times 768$ (Figure~\ref{fig:mlwe_feature_extraction}-center-left), as suggested by \cite{ethayarajh2019contextual}.

Since BERT works with word pieces but \XXX\ objective is to extract full tokens (words), the embedding of word pieces belonging to the same word are aggregated, averaging their values over the word-piece axes, and obtaining a new matrix of tokens embedding of shape $t \times 768$, where $t$ is the number of input tokens (Figure~\ref{fig:mlwe_feature_extraction}-center-right).
In the end, due to the sparse nature of data, the dimensionality reduction technique, i.e., Principal Component Analysis, is exploited, reducing the final shape of the tokens embeddings matrix to $(t \times c)$, where $c$ is the number of extracted components (Figure~\ref{fig:mlwe_feature_extraction}-right).
This last result is the \textit{Multi-layer word embedding} representation for the BERT model.

\subsection{Experiments at a glance} 

For each input document, the local explanations were computed exploiting all the feature extraction methods described in Section~\ref{sec:architecture}, for both use cases.
\begin{table}
        \centering
        \small
        \begin{tabular}{|c|c|c|}
            \hline
            Feature extraction type & No combination & Pairwise combination \\ \hline
             Part-of-speech & 33\% & 70\%  \\ \hline
             Sentence & 22\% & 30\%  \\ \hline
             MLWE & 75\% & 86\% \\ \hline
             Overall & 80\% & 90\% \\ \hline
        \end{tabular}
        \captionof{table}{Explanations of the BERT model: percentage of documents for which each feature extraction strategy produces at least one informative local explanation (i.e., with $nPIR \geqslant 0.5$), with and without combination of features. \textit{Overall} is the percentage of documents for which at least one method provided a local explanation with $nPIR \geqslant 0.5$.}
        \label{tab:overall-experiments}
\end{table}

In the explanation process of the sentiment analysis use case with the BERT model, \XXX\ has been experimentally evaluated on 400 textual documents, 202 belonging to the class \textit{Positive} and 198 to the class \textit{Negative}, for a total of almost 100,000 local explanations, with an average of 250 local explanations for each input document.
However, only the most informative local explanations are automatically shown by the engine to the user.
A local explanation has been defined to be informative when having a $nPIR$ value equal to or higher than 0.5.
All the rest of the local explanations produced by \XXX\ are still available to the users, should they liked to investigate further insights into the prediction process.
To show the effectiveness of the proposed feature extraction techniques, we analyzed the percentage of documents for which \XXX\ computed local explanations with at least one informative feature for the class-of-interest. 
Experiments on the same input texts have been repeated twice, firstly without combining the different features, then including the pairwise combinations for each feature extraction method. 
Table~\ref{tab:overall-experiments} shows the percentage of documents required to find at least one informative feature ($nPIR \geqslant 0.5$) with and without combinations of pairwise features. 
The MLWE method leads to abundantly better results than the other methods. 
The part-of-speech strategy benefits the most from the pairwise combinations, allowing to create features representing more complex concepts. 
For example, the combination of \textit{adjectives} and \textit{nouns} allows to create features composed by words like \{{\fontfamily{cmtt}\selectfont bad, film}\} that, together, can better express a sentiment.

\begin{table}
        \centering
        \small
        \begin{tabular}{|c|c|c|c|}
            \hline
             Feature Extraction Type & Clean & Toxic & Clean/Toxic \\ \hline
             Part-of-speech & 8\% & 98\% & 53\%  \\ \hline
             Sentence & 2\% & 76\% & 39\%  \\ \hline
             MLWE & 12\% & 98\% & 55\% \\ \hline
             Overall & 15\% & 99\% & 58\% \\ \hline
        \end{tabular}
        \captionof{table}{Explanation of the custom LSTM model: percentage of documents for which each feature extraction strategy produces at least one informative local explanation (i.e. with $nPIR \geqslant 0.5$), with combination of features, for the class labels \textit{Clean} and \textit{Toxic}. 
         }
        \label{tab:overall-experiments-toxic}
\end{table}

In the explanation process of the toxic comment use case with the custom LSTM model, \XXX\ has been experimentally evaluated on 2250 documents, 1121 belonging to the class \textit{Toxic} and 1129 to class \textit{Clean}, leading to almost 160,000 local explanations in total.
Table~\ref{tab:overall-experiments-toxic} shows the percentage of input documents for which \XXX\ has been able to extract at least one informative local explanation ($nPIR \geqslant 0.5$).
For the \textit{Toxic} class, \XXX\ has been able to identify at least one informative explanation for almost all the documents, with most of the feature extraction strategies. 
Only the sentence-based feature extraction has a lower percentage of informative explanations w.r.t. the other techniques. This suggests that toxic words tend to be sparse in the input text and not concentrated in a single sentence.
Finding informative explanations for the \textit{Clean} class has proven to be harder.
None of the feature extraction techniques can explain more than 15\% of the predictions for the \textit{Clean} input texts.
The nature of the use case under exam can explain possible causes: usually, a document is considered clean; it can become toxic because of the presence of specific words or linguistic expressions.
Thus, the hypothesis is that there is no specific pattern of words that represents the \textit{Clean} class (see Section~\ref{ssec:global-insights}
for further details).

\subsection{Local explanation} 
The purpose of this Section is to discuss some specific local explanations provided by \XXX\ in different conditions, to show their relevance and usefulness in explaining the deep NLP model behavior, for both the custom LSTM and the BERT models of the two use cases.

\begin{figure}
    \centering
    \captionsetup[subfloat]{farskip=6pt,captionskip=0pt}
    \subfloat[Original text]{
        \label{fig:ex-4-original}
        \fbox{\begin{minipage}{0.9\columnwidth}
        \scriptsize
        \linespread{1}
        \fontfamily{cmtt}\selectfont
        Criticize a black man and the left calls you a racist. Criticize a woman and you are a sexist. Now I will criticize you as a fool and you can call me intolerant.
        \end{minipage}}
    }\\
    \subfloat[\texttt{EXP1:} Adjective \& Noun - POS feature extraction]{
        \label{fig:ex-4-adj-noun}
        \fbox{\begin{minipage}{0.9\columnwidth}
        \scriptsize
        \fontfamily{cmtt}\selectfont
        Criticize a \textcolor{ansi-red-intense}{black man} and the \textcolor{ansi-red-intense}{left} calls you a \textcolor{ansi-red-intense}{racist}. Criticize a \textcolor{ansi-red-intense}{woman} and you are a \textcolor{ansi-red-intense}{sexist}. Now I will criticize you as a \textcolor{ansi-red-intense}{fool} and you can call me \textcolor{ansi-red-intense}{intolerant}.
        \end{minipage}}
    }\\
   
    \subfloat[\texttt{EXP2:} Multi-layer word embedding feature extraction]{
        \label{fig:ex-4-mlwe}
        \fbox{\begin{minipage}{0.9\columnwidth}
        \scriptsize
        \fontfamily{cmtt}\selectfont
        Criticize a \textcolor{ansi-red-intense}{black man} and the \textcolor{ansi-red-intense}{left} calls you a \textcolor{ansi-red-intense}{racist}. Criticize a \textcolor{ansi-red-intense}{woman} and you are a \textcolor{ansi-red-intense}{sexist}. \textcolor{ansi-red-intense}{Now} I will criticize you as a \textcolor{ansi-red-intense}{fool} and you can call me \textcolor{ansi-red-intense}{intolerant}.
        \end{minipage}}
    }
    
    \caption{Examples of \textit{textual explanation} report for the input in Figure~\ref{fig:ex-4-original} originally labeled by custom LSTM model as \textit{Toxic} with a probability of 0.98. The most relevant features are highlighted in red.}
    \label{fig:ex-4-textual-explanation}
    
    
    \begin{minipage}{\columnwidth}
        \centering
        \small
        \begin{tabular}{|c|c|c|c|c|}
            \hline
            Explanation & Feature $f$ & $L_o$ & $L_f$  & $nPIR_f(N)$ \\ \hline
             EXP1 & POS-Adj\&Noun & T & C & 0.839 \\ \hline
             EXP2 & MLWE & T & C & 0.883 \\ \hline

        \end{tabular}
        \captionof{table}{Quantitative explanation for the example reported in Figure~\ref{fig:ex-4-textual-explanation}. T is the \textit{Toxic} label, C is the \textit{Clean} label.}
        \label{tab:ex-4-numerical-explanation}
    \end{minipage}
\end{figure}

\vspace{0.2cm}
\noindent
\textbf{Local Explanation 1.} 
In the first example, reported in Figure~\ref{fig:ex-4-textual-explanation}, the custom LSTM model classifies the input comment in Figure~\ref{fig:ex-4-original} as \textit{Toxic}. 
The most influential features identified by \XXX\ are shown in Figures~\ref{fig:ex-4-adj-noun} and \ref{fig:ex-4-mlwe}. 
The different feature extraction strategies find that the most positively influential features for the \textit{Toxic} class labels are \{{\fontfamily{cmtt}\selectfont black man, left, racist, woman, sexist, fool, intolerant}\}. 
In particular, the most informative explanations are extracted with the combination of adjectives and nouns (Table~\ref{tab:ex-4-numerical-explanation}-EXP1) and with MLWE (Table~\ref{tab:ex-4-numerical-explanation}-EXP2). 
It is interesting to notice that in this case, the combination of \textit{adjectives} and \textit{nouns} is very relevant for this model, e.g., it is not just the word \texttt{black} that makes a comment toxic, but the combination \texttt{black man}.
Furthermore, the POS feature extraction and the MLWE highlighted very similar sets of words. 
In this case, the prediction is trustful, and in particular, it is relevant that the model learned features like \texttt{black man} and \texttt{woman} to be influential for the \textit{Toxic} class.

\begin{figure}
    \centering
    \scriptsize
    \captionsetup[subfloat]{farskip=6pt,captionskip=0pt}
    \subfloat[Original text]{
        \label{fig:ex-1-original}
        \fbox{\begin{minipage}{0.9\columnwidth}
        \linespread{1}
        \scriptsize
        \fontfamily{cmtt}\selectfont
        How many movies are there that you can think of when you see a movie like this?
        I can't count them but it sure seemed like the movie makers were trying to give
        me a hint. I was reminded so often of other movies, it became a big distraction.
        One of the borrowed memorable lines came from a movie from 2003 - Day After
        Tomorrow. One line by itself, is not so bad but this movie borrows so much from
        so many movies it becomes a bad risk.
        BUT{\ldots}
        See The Movie! Despite its downfalls there is enough to make it interesting and
        maybe make it appear clever. While borrowing so much from other movies it never
        goes overboard. In fact, you'll probably find yourself battening down the
        hatches and riding the storm out. Why? {\ldots}Costner and Kutcher played their
        characters very well. I have never been a fan of Kutcher's and I nearly gave up
        on him in The Guardian, but he surfaced in good fashion. Costner carries the
        movie swimmingly with the best of Costner's ability. I don't think Mrs. Robinson
        had anything to do with his success.
        The supporting cast all around played their parts well. I had no problem with
        any of them in the end. But some of these characters were used too much.
        From here on out I can only nit-pick so I will save you the wear and tear. Enjoy
        the movie, the parts that work, work well enough to keep your head above water.
        Just don't expect a smooth ride.
        7 of 10 but almost a 6.
        \end{minipage}}
    }\\
    \subfloat[\texttt{EXP1:} Adjective - POS feature extraction]{
        \label{fig:ex-1-adjective}
        \fbox{\begin{minipage}{0.9\columnwidth}
        \scriptsize
        \fontfamily{cmtt}\selectfont
        How \textcolor{ansi-red-intense}{many} movies are there that you can think of when you see a movie
        [...]
        I was reminded so often of \textcolor{ansi-red-intense}{other} movies, it
        became a \textcolor{ansi-red-intense}{big} distraction. One of the borrowed \textcolor{ansi-red-intense}{memorable} lines
        came from a movie from 2003 - Day After Tomorrow. One line by itself, is not so \textcolor{ansi-red-intense}{bad} but this movie borrows so \textcolor{ansi-red-intense}{much} from so \textcolor{ansi-red-intense}{many}
        movies it becomes a \textcolor{ansi-red-intense}{bad} risk. BUT {\ldots} See The Movie! Despite its
        downfalls there is \textcolor{ansi-red-intense}{enough} to make it \textcolor{ansi-red-intense}{interesting} and maybe
        make it appear clever. While borrowing so much from \textcolor{ansi-red-intense}{other} movies it
        never goes overboard. 
        [...]
        I have never been a fan of Kutcher 's and I nearly gave up
        on him in The Guardian, but he surfaced in \textcolor{ansi-red-intense}{good} fashion. Costner
        carries the movie swimmingly with the \textcolor{ansi-red-intense}{best} of Costner 's ability. 
        [...] But some of these characters were used too \textcolor{ansi-red-intense}{much}. [...]
        Just do n't expect a \textcolor{ansi-red-intense}{smooth} ride. 7 of 10 but almost a 6.
        \end{minipage}}
    }\\
    \subfloat[\texttt{EXP2:} Sentence feature extraction]{
        \label{fig:ex-1-sentence}
        \fbox{\begin{minipage}{0.9\columnwidth}
        \scriptsize
        \fontfamily{cmtt}\selectfont
        How many movies are there that you can think of when you see a movie like this?
        I can't count them but it sure seemed like the movie makers were trying to give
        me a hint. \textcolor{ansi-red-intense}{I} \textcolor{ansi-red-intense}{was} \textcolor{ansi-red-intense}{reminded} \textcolor{ansi-red-intense}{so} \textcolor{ansi-red-intense}{often}
        \textcolor{ansi-red-intense}{of} \textcolor{ansi-red-intense}{other} \textcolor{ansi-red-intense}{movies}\textcolor{ansi-red-intense}{,} \textcolor{ansi-red-intense}{it} \textcolor{ansi-red-intense}{became} \textcolor{ansi-red-intense}{a} \textcolor{ansi-red-intense}{big} \textcolor{ansi-red-intense}{distraction}\textcolor{ansi-red-intense}{.} One of [...] 
        \end{minipage}}
    }\\
    \subfloat[\texttt{EXP3:} Multi-layer word embedding feature extraction]{
        \label{fig:ex-1-mlwe}
        \fbox{\begin{minipage}{0.9\columnwidth}
        \scriptsize
        \fontfamily{cmtt}\selectfont
        How many movies are \textcolor{ansi-red-intense}{there} that you can think of when you see a movie like this? 
        [...] 
        See the movie despite its downfalls \textcolor{ansi-red-intense}{there} is enough to make it interesting and maybe make it appear clever.
        [...] 
        \end{minipage}}
    }
    
    \caption{Examples of \textit{textual explanation} report for the input in Figure~\ref{fig:ex-1-original}, wrongly labeled by BERT as \textit{Negative} with a probability of 0.99. The most relevant features are highlighted in red.}
    \label{fig:ex-1-textual-explanation}
    
    
    \begin{minipage}{\columnwidth}
        \centering
        \small
        \begin{tabular}{|c|c|c|c|c|}
            \hline
            Explanation & Feature $f$ & $L_o$ & $L_f$  & $nPIR_f(N)$ \\ \hline
             EXP1 & POS-Adjective & N & P & 0.884 \\ \hline
             EXP2 & Sentence & N & P & 0.663 \\ \hline
             EXP3 & MLWE & N & P & 0.651 \\ \hline
        \end{tabular}
        \captionof{table}{Quantitative explanation for the example in Figure~\ref{fig:ex-1-textual-explanation}. P is the positive label, N is the negative label.}
        \label{tab:ex-1-numerical-explanation}
    \end{minipage}
\end{figure}

\vspace{0.2cm}
\noindent
\textbf{Local Explanation 2.}
In the second example, the BERT model makes a wrong prediction by classifying the sentiment of the input text in Figure \ref{fig:ex-1-original} as \textit{Negative}, while the expected label (ground-truth) is \textit{Positive}.
A user requiring to decide whether to trust such prediction can take advantage of \XXX\ to understand which are the words influencing the outcome.
Figure~\ref{fig:ex-1-textual-explanation} shows the \textit{textual explanations} provided by the most influential features.
Table~\ref{tab:ex-1-numerical-explanation} contains the corresponding \textit{quantitative explanations} with the $nPIR$ values.
\XXX\ identified three local explanations for the \textit{Negative} class with $nPIR$ values higher than 0.5, whose perturbation would cause a change in the predicted label from \textit{Negative} to \textit{Positive}).
The top informative features were extracted exploiting Adjectives-POS (Figure~\ref{fig:ex-1-adjective}), Sentence (Figure~\ref{fig:ex-1-sentence}) and MLWE (Figure~\ref{fig:ex-1-mlwe}).
Regarding the Adjectives-POS feature extraction, Figure~\ref{fig:ex-1-adjective} shows that general words like \{{\fontfamily{cmtt}\selectfont many, other, big, ..., smooth}\} have a $nPIR$ value for the class \textit{Negative} close to 0.88 (Table~\ref{tab:ex-1-numerical-explanation}-EXP1). 
General words with a very strong impact on the final prediction for this specific input text is not a trustful indicator: their absence might lead to entirely different outcomes.\\
Regarding the sentence-based feature extraction, the negative prediction is triggered by only one specific phrase (Figure~\ref{fig:ex-1-sentence}), whose absence leads to a \textit{Positive} prediction with a $nPIR$ value of $0.66$ (Table~\ref{tab:ex-1-numerical-explanation}-EXP2).\\
Finally, the MLWE feature extraction strategy identifies a cluster composed by only two instances of a single very general word, \{{\fontfamily{cmtt}\selectfont there}\} (Figure~\ref{fig:ex-1-mlwe}).
By removing the two occurrences of the word \{{\fontfamily{cmtt}\selectfont there}\}, the prediction changes from \textit{Negative} to \textit{Positive} with a $nPIR$ value of 0.651 (Table~\ref{tab:ex-1-numerical-explanation}-EXP3).\\
Since the output of the prediction model can be drifted (from \textit{Negative} to \textit{Positive}) by simply removing occurrences of general words such as \{{\fontfamily{cmtt}\selectfont there, many, other, big, smooth, ...}\} from the input text (actually removing only \{{\fontfamily{cmtt}\selectfont there}\} is enough!), doubts on the predicted class reliability are reasonable.
More details related to the global behavior and the robustness of the model are addressed in Section~\ref{ssec:global-insights}.


\begin{figure}
    \centering
    \captionsetup[subfloat]{farskip=6pt,captionskip=0pt}
    \subfloat[Original text]{
        \label{fig:ex-2-original}
        \fbox{\begin{minipage}{0.9\columnwidth}
        \scriptsize
        \linespread{1}
        \fontfamily{cmtt}\selectfont
        There were so many classic movies that were made where the leading people were out-and- out liars and yet they are made to look good. I never bought into that stuff. The "screwball comedies" were full of that stuff and so were a lot of the Fred Astaire films. Here, Barbara Stanwyck plays a famous "country" magazine writer who has been lying to the public for years, and feels she has to keep lying to keep her persona (and her job). She even lies to a guy about getting married, another topic that was always trivialized in classic films. She's a New York City woman who pretends she's a great cook and someone who knows how to handle babies, etc. Obviously she knows nothing and the lies pile up so fast you lose track. I guess all of that is supposed to be funny because lessons are learned in the end and true love prevails, etc. etc. Please pass the barf bag. Most of this film is NOT funny. Stanwyck was far better in the film noir genre. As for Dennis Morgan, well, pass the bag again.
        \end{minipage}}
    }\\
    \subfloat[\texttt{EXP1:} Adjective - POS feature extraction]{
        \label{fig:ex-2-adjective}
        \fbox{\begin{minipage}{0.9\columnwidth}
        \scriptsize
        \fontfamily{cmtt}\selectfont
        There were so \textcolor{ansi-red-intense}{many} \textcolor{ansi-red-intense}{classic} movies that were made where the leading people were \textcolor{ansi-red-intense}{out-and-} out liars and yet they are made to look \textcolor{ansi-red-intense}{good}. I never bought into that stuff. The ``\textcolor{ansi-red-intense}{screwball} comedies'' were \textcolor{ansi-red-intense}{full} of 
        [...] plays a \textcolor{ansi-red-intense}{famous} ``country'' 
        [...] getting
        \textcolor{ansi-red-intense}{married}, another topic that was always trivialized in \textcolor{ansi-red-intense}{classic} films.
        [...] she's a \textcolor{ansi-red-intense}{great} cook and someone
        [...] supposed to be \textcolor{ansi-red-intense}{funny} because lessons are learned in the end and \textcolor{ansi-red-intense}{true} love
        [...] bag. \textcolor{ansi-red-intense}{Most} of this film is NOT \textcolor{ansi-red-intense}{funny}. Stanwyck was far better in the film \textcolor{ansi-red-intense}{noir} genre. [...]
        \end{minipage}}
    }\\

    \subfloat[\texttt{EXP2:} Verb - POS feature extraction]{
        \label{fig:ex-2-verb}
        \fbox{\begin{minipage}{0.9\columnwidth}
        \scriptsize
        \fontfamily{cmtt}\selectfont
        There \textcolor{ansi-red-intense}{were} so 
        [...] that \textcolor{ansi-red-intense}{were} \textcolor{ansi-red-intense}{made}
        where the \textcolor{ansi-red-intense}{leading} people \textcolor{ansi-red-intense}{were} out-and- out liars and yet they
        \textcolor{ansi-red-intense}{are} \textcolor{ansi-red-intense}{made} to \textcolor{ansi-red-intense}{look} good. I never \textcolor{ansi-red-intense}{bought} into
        that stuff. The ``screwball comedies'' \textcolor{ansi-red-intense}{were} full of that stuff and so
        \textcolor{ansi-red-intense}{were} a lot [...] 
        Barbara Stanwyck
        \textcolor{ansi-red-intense}{plays} a famous ``country'' magazine writer who \textcolor{ansi-red-intense}{has}
        \textcolor{ansi-red-intense}{been} \textcolor{ansi-red-intense}{lying} to
        [...] she \textcolor{ansi-red-intense}{has}
        to \textcolor{ansi-red-intense}{keep} \textcolor{ansi-red-intense}{lying} to \textcolor{ansi-red-intense}{keep} her persona (and her job).
        She even \textcolor{ansi-red-intense}{lies} to a guy about \textcolor{ansi-red-intense}{getting} married, another topic
        that \textcolor{ansi-red-intense}{was} always \textcolor{ansi-red-intense}{trivialized} in classic films. She \textcolor{ansi-red-intense}{'s}
        a New York City woman who \textcolor{ansi-red-intense}{pretends} she\textcolor{ansi-red-intense}{'s} a great cook and
        someone who \textcolor{ansi-red-intense}{knows} how to \textcolor{ansi-red-intense}{handle} babies, etc. Obviously she
        \textcolor{ansi-red-intense}{knows} nothing and the lies \textcolor{ansi-red-intense}{pile} \textcolor{ansi-red-intense}{up} so fast you
        \textcolor{ansi-red-intense}{lose} track. I \textcolor{ansi-red-intense}{guess} all of that \textcolor{ansi-red-intense}{is} \textcolor{ansi-red-intense}{supposed}
        to \textcolor{ansi-red-intense}{be} funny because lessons \textcolor{ansi-red-intense}{are} \textcolor{ansi-red-intense}{learned} in the [...]
        Please \textcolor{ansi-red-intense}{pass} the barf bag. Most of
        this film \textcolor{ansi-red-intense}{is} NOT funny. Stanwyck \textcolor{ansi-red-intense}{was} far 
        [...] well, \textcolor{ansi-red-intense}{pass} the bag again.
        \end{minipage}}
    }\\
    \subfloat[\texttt{EXP3:} Adjective \& Verb - POS feature extraction]{
        \label{fig:ex-2-adjective-verb}
        \fbox{\begin{minipage}{0.9\columnwidth}
        \scriptsize
        \fontfamily{cmtt}\selectfont
        There \textcolor{ansi-red-intense}{were} so \textcolor{ansi-red-intense}{many} \textcolor{ansi-red-intense}{classic} movies that \textcolor{ansi-red-intense}{were}
        \textcolor{ansi-red-intense}{made} where the \textcolor{ansi-red-intense}{leading} people \textcolor{ansi-red-intense}{were} \textcolor{ansi-red-intense}{out-and-}
        out liars and yet they \textcolor{ansi-red-intense}{are} \textcolor{ansi-red-intense}{made} to \textcolor{ansi-red-intense}{look}
        \textcolor{ansi-red-intense}{good}. I never \textcolor{ansi-red-intense}{bought} into that stuff. The ``\textcolor{ansi-red-intense}{screwball} comedies'' \textcolor{ansi-red-intense}{were} \textcolor{ansi-red-intense}{full} of that stuff and so
        \textcolor{ansi-red-intense}{were} a lot of the Fred Astaire films. Here, Barbara Stanwyck
        \textcolor{ansi-red-intense}{plays} a \textcolor{ansi-red-intense}{famous} ``country'' magazine writer who \textcolor{ansi-red-intense}{has}
        \textcolor{ansi-red-intense}{been} \textcolor{ansi-red-intense}{lying} to the public for years, and feels she \textcolor{ansi-red-intense}{has}
        to \textcolor{ansi-red-intense}{keep} \textcolor{ansi-red-intense}{lying} to \textcolor{ansi-red-intense}{keep} her persona (and her job).
        She even \textcolor{ansi-red-intense}{lies} to a guy about \textcolor{ansi-red-intense}{getting} \textcolor{ansi-red-intense}{married}, another
        topic that \textcolor{ansi-red-intense}{was} always \textcolor{ansi-red-intense}{trivialized} in \textcolor{ansi-red-intense}{classic} films.
        She \textcolor{ansi-red-intense}{'s} a New York City woman who \textcolor{ansi-red-intense}{pretends} she \textcolor{ansi-red-intense}{'s} a
        \textcolor{ansi-red-intense}{great} cook and someone who \textcolor{ansi-red-intense}{knows} how to \textcolor{ansi-red-intense}{handle}
        babies, etc. Obviously she \textcolor{ansi-red-intense}{knows} nothing and the lies \textcolor{ansi-red-intense}{pile}
        \textcolor{ansi-red-intense}{up} so fast you \textcolor{ansi-red-intense}{lose} track. I \textcolor{ansi-red-intense}{guess} all of that
        \textcolor{ansi-red-intense}{is} \textcolor{ansi-red-intense}{supposed} to \textcolor{ansi-red-intense}{be} \textcolor{ansi-red-intense}{funny} because lessons
        \textcolor{ansi-red-intense}{are} \textcolor{ansi-red-intense}{learned} in the end and \textcolor{ansi-red-intense}{true} love prevails, etc.
        etc. Please \textcolor{ansi-red-intense}{pass} the barf bag. \textcolor{ansi-red-intense}{Most} of this film \textcolor{ansi-red-intense}{is}
        NOT \textcolor{ansi-red-intense}{funny}. Stanwyck \textcolor{ansi-red-intense}{was} far better in the film \textcolor{ansi-red-intense}{noir}
        genre. As for Dennis Morgan, well, \textcolor{ansi-red-intense}{pass} the bag again.
        \end{minipage}}
    }
    
   \caption{Examples of \textit{textual explanation} report for the input in Figure~\ref{fig:ex-2-original} originally labeled by BERT as \textit{Negative} with a probability of 0.99. Features found are highlighted in red.(Continue)}
    \label{fig:ex-2-textual-explanation}
\end{figure}

\begin{figure}
    \centering
    \ContinuedFloat
    \subfloat[\texttt{EXP4:} Sentence feature extraction]{
        \label{fig:ex-2-sentence}
        \fbox{\begin{minipage}{0.9\columnwidth}
        \scriptsize
        \fontfamily{cmtt}\selectfont
        [...] \textcolor{ansi-red-intense}{She} \textcolor{ansi-red-intense}{even}
        \textcolor{ansi-red-intense}{lies} \textcolor{ansi-red-intense}{to} \textcolor{ansi-red-intense}{a} \textcolor{ansi-red-intense}{guy} \textcolor{ansi-red-intense}{about}
        \textcolor{ansi-red-intense}{getting} \textcolor{ansi-red-intense}{married}\textcolor{ansi-red-intense}{,} \textcolor{ansi-red-intense}{another} \textcolor{ansi-red-intense}{topic}
        \textcolor{ansi-red-intense}{that} \textcolor{ansi-red-intense}{was} \textcolor{ansi-red-intense}{always} \textcolor{ansi-red-intense}{trivialized} \textcolor{ansi-red-intense}{in}
        \textcolor{ansi-red-intense}{classic} \textcolor{ansi-red-intense}{films}\textcolor{ansi-red-intense}{.} [...]
        \end{minipage}}
    }\\

    \subfloat[\texttt{EXP5:} Multi-layer word embedding feature extraction]{
        \label{fig:ex-2-mlwe}
        \fbox{\begin{minipage}{0.9\columnwidth}
        \scriptsize
        \fontfamily{cmtt}\selectfont
        [...] \textcolor{ansi-red-intense}{I} never
        bought into that \textcolor{ansi-red-intense}{stuff}. The "screwball comedies" were \textcolor{ansi-red-intense}{full}
        of that stuff and \textcolor{ansi-red-intense}{so} were a lot of the \textcolor{ansi-red-intense}{Fred} Astaire films.
        Here, Barbara Stanwyck plays a famous "country" magazine writer who has
        \textcolor{ansi-red-intense}{been} lying to the public for years, and feels she has to keep lying to
        keep her persona (and her job). she even lies to a guy about getting
        married, \textcolor{ansi-red-intense}{another} topic that was always \textcolor{ansi-red-intense}{trivialized} in classic
        films. she's a new york city woman who \textcolor{ansi-red-intense}{pretends} \textcolor{ansi-red-intense}{she}'\textcolor{ansi-red-intense}{s} a great cook and someone who \textcolor{ansi-red-intense}{knows} how to handle babies,
        etc. Obviously she knows nothing and the lies pile up \textcolor{ansi-red-intense}{so} fast you lose
        track. \textcolor{ansi-red-intense}{I} \textcolor{ansi-red-intense}{guess} all of that is supposed to \textcolor{ansi-red-intense}{be}
        \textcolor{ansi-red-intense}{funny} because lessons are learned in the end and true love prevails,
        etc. [...] Most \textcolor{ansi-red-intense}{of} this film is not funny.
        Stanwyck was far better in the film noir genre. as for Dennis Morgan,
        \textcolor{ansi-red-intense}{well}, \textcolor{ansi-red-intense}{pass} the bag again
        \end{minipage}}
    }\\

    \caption{(Continued) Examples of \textit{textual explanations} for the input in Figure~\ref{fig:ex-2-original}, originally labeled by BERT as \textit{Negative} with a probability of 0.99. Features extracted by \XXX\ are highlighted in red.}
    \label{fig:ex-2-textual-explanation}
    
    
    \begin{minipage}{\columnwidth}
        \centering
        \small
        \begin{tabular}{|c|c|c|c|c|}
            \hline
            Explanation & Feature $f$ & $L_o$ & $L_f$  & $nPIR_f(N)$ \\ \hline
             EXP1 & POS-Adjective & N & N & 0.003 \\ \hline
             EXP2 & POS-Verb & N & N & 0.137 \\ \hline
             EXP3 & POS-Adj\&Verb & N & P & 0.915 \\ \hline
             EXP4 & Sentence & N & P & 0.638 \\ \hline
             EXP5 & MLWE & N & P & 0.899 \\ \hline
        \end{tabular}
        \captionof{table}{Quantitative explanations for the example reported in Figure~\ref{fig:ex-2-textual-explanation}. P is the positive label, N is the negative label.}
        \label{tab:ex-2-numerical-explanation}
    \end{minipage}
\end{figure}

\vspace{0.2cm}
\noindent
\textbf{Local Explanation 3.} 
The example is reported in Figure~\ref{fig:ex-2-textual-explanation}, where the BERT model correctly classifies the input text in Figure~\ref{fig:ex-2-original} as \textit{Negative}. 
The \textit{textual explanations} produced by \XXX\ exploiting different feature extraction strategies are reported as follows:
adjective-POS in Figure~\ref{fig:ex-2-adjective},
verb-POS in Figure~\ref{fig:ex-2-verb},
adjective-verb-POS in Figure~\ref{fig:ex-2-adjective-verb}, 
sentence in Figure~\ref{fig:ex-2-sentence}, and 
multi-layer word embedding in Figure~\ref{fig:ex-2-mlwe}.
Their $nPIR$ values are reported in the \textit{quantitative explanations} of Table~\ref{tab:ex-2-numerical-explanation}.
We note that only the adjective-verb-POS, sentence, and MLWE techniques provide informative explanations, whereas the adjective-POS and verb-POS yield uninformative explanations, yet we include them in the example for discussion.

The POS feature analysis (Figures~\ref{fig:ex-2-adjective},~\ref{fig:ex-2-verb}) shows that the different parts-of-speech, taken separately one at a time, are not influential for the prediction of the class \textit{Negative}. 
From the \textit{quantitative explanation} of EXP1 and EXP2 in Table \ref{tab:ex-2-numerical-explanation} indeed it can be observed that they achieve a \textit{nPIR} close to 0.003 and 0.137 respectively. A similar result was obtained for all the other POS features considered individually.
Consequently, \XXX\ explores the \textit{pairwise combinations} (as explained in Section \ref{ssec:interpretable-feature-exctraction}) of the parts-of-speech to create more sophisticated features and to analyze more complex semantic concepts. 
In this case, the feature composed by \textit{Adjectives} and \textit{Verbs} (Figure \ref{fig:ex-2-adjective-verb}) is reported to be impacting for the predicted class label reaching a \textit{nPIR} value close to 0.915 (EXP3 in Table \ref{tab:ex-2-numerical-explanation}).\\
The sentence feature extraction strategy, instead, identifies the feature composed by the phrase in Figure \ref{fig:ex-2-sentence} as positively influential for the predicted class with a \textit{nPIR} score of about 0.638 (EXP4 in Table \ref{tab:ex-2-numerical-explanation}).\\
Finally, the MLWE feature extraction strategy identifies $K=15$ as the most influential partitioning of words. 
Analyzing the 15 different features, composed by clusters of words, emerges that the only one with a significant impact on the output prediction is that showed in Figure \ref{fig:ex-2-mlwe}, reaching a \textit{nPIR} of 0.899 (EXP5 in Table \ref{tab:ex-2-numerical-explanation}).\\
Analyzing the content of the most informative textual explanations (EXP3, EXP4 and EXP5), it can be observed that, interestingly, all the local explanations with high values of $nPIR$ contain the word \{{\fontfamily{cmtt}\selectfont trivialized}\}. It might seem that a single word can be the only responsible for the original prediction. However, also the explanation EXP2 contains the same word but is not influential for the class label. Therefore, it emerges that the output predictions are not influenced by single words, but is the combination of different words that allows creating more complex concepts which determine the predicted class label.
Moreover, it is possible to say that, in this specific prediction, the model is not sensible to the perturbation of adjectives (EXP1 in figure~\ref{fig:ex-2-adjective}) or verbs (EXP2 in Figure~\ref{fig:ex-2-verb}) separately, highlighting that the proposed prediction has been produced taking into account the whole context of the input text.
Only in EXP3 (Figure~\ref{fig:ex-2-adjective-verb}), it is possible to notice that, when adjectives and verbs are perturbed together, changing the meaning of the input text radically, the predicted class changes.
The joint perturbation can be considered a good measure of robustness for the prediction performed by the fine-tuned BERT model under analysis.\\
However, as for the previous example, it is shown in EXP4 (Figure~\ref{fig:ex-2-sentence}) that exist a singular phrase more relevant than the others in the decision-making process.
The perturbation of the sentence in EXP4 will bring the model to change the prediction from class \textit{Negative} to \textit{Positive}.\\
Furthermore, EXP5 (Figure~\ref{fig:ex-2-mlwe}), obtained through the MLWE feature extraction technique, shows an apparently random pool of words very relevant in the prediction process.
The MLWE feature extraction is able to find the influential feature with higher precision concerning EXP3 (obtained by the combination of all verbs and adjectives), with a very small penalty on the nPIR score.
Indeed, the MLWE strategy is able to find a small number of words belonging to different part-of-speeches and different sentences that are affecting the model's output. So, also the resulting explanations are more understandable and meaningful for the end-user.

As in the previous example, this last experiment shows that the predictive model is particularly sensitive to a few specific variations of, apparently not correlated, input words.

\vspace{0.2cm}
From these examples, it emerges that the different feature extraction strategies should be used in a complementary manner, as they look at different aspects of the input text and provide different kinds of explanations. Furthermore, the proposed examples showed that: 
\begin{itemize}
\item 
    \XXX\ can be successfully applied to different deep learning models;
\item 
    the proposed prediction explanation process can be applied with success to different use cases and NLP tasks;
\item 
    \XXX\ can extract meaningful explanations from both long and short text documents without limiting their interpretability;
\item 
    the end-user is provided with informative details to analyze critically and judge the quality of the model outcomes, being supported in deciding whether its decision-making process is trustful.
\end{itemize}

\subsection{Model-global explanations}
\label{ssec:global-insights}
Exploiting the prediction-local explanations computed by \XXX\ for all the input documents, \textit{model-global} insights can be provided.

\begin{figure*}[h!]%
    \centering
    \subfloat[GAI \textit{Toxic} class.]{{
    \includegraphics[width=0.4\textwidth]{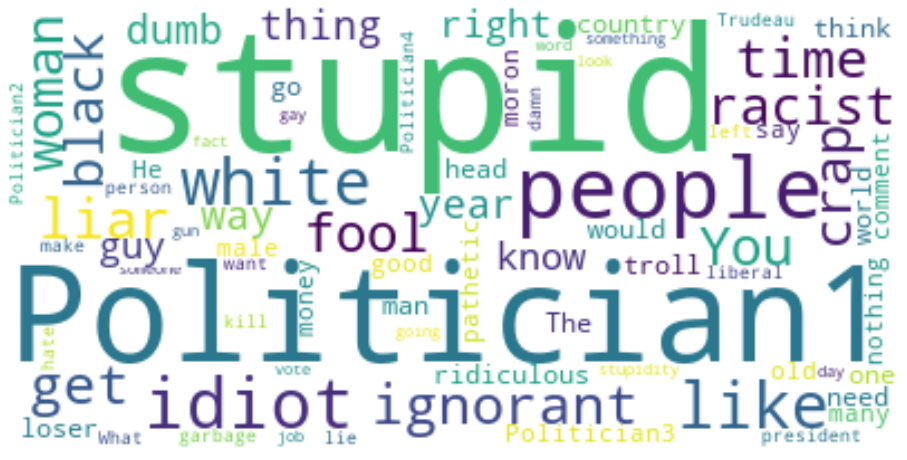}\label{fig:wordcloud-toxic-gai}}}%
    \qquad
    \subfloat[GAI \textit{Clean} class.]{{
    \includegraphics[width=0.4\textwidth]{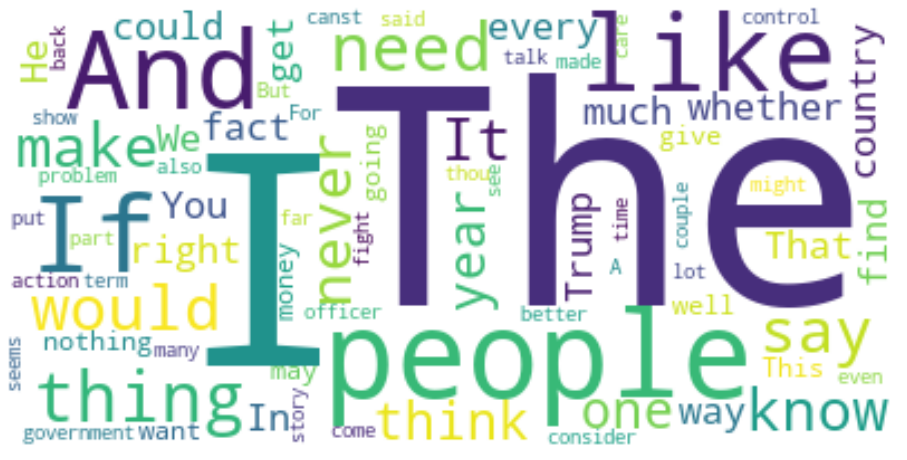}\label{fig:wordcloud-clean-gai}}}%
    
    \subfloat[GRI \textit{Toxic} class.]{{
    \includegraphics[width=0.4\textwidth]{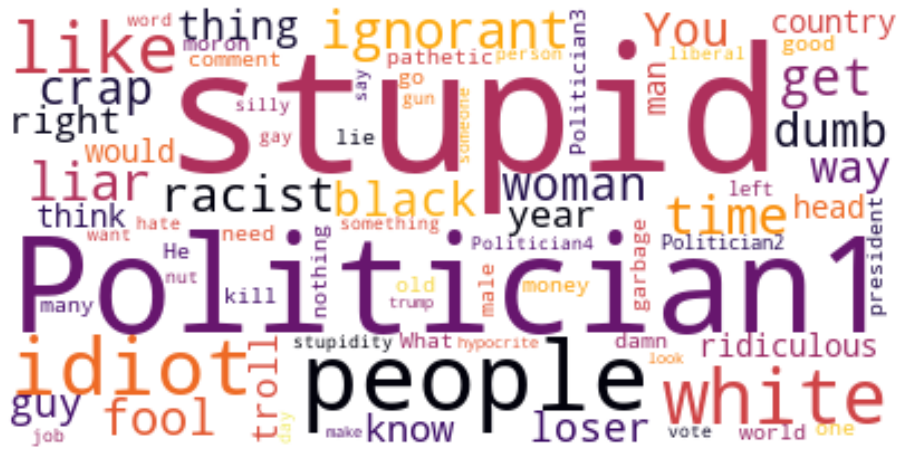}\label{fig:wordcloud-toxic-gri}}}%
    \qquad
    \subfloat[GRI \textit{Clean} class.]{{
    \includegraphics[width=0.4\textwidth]{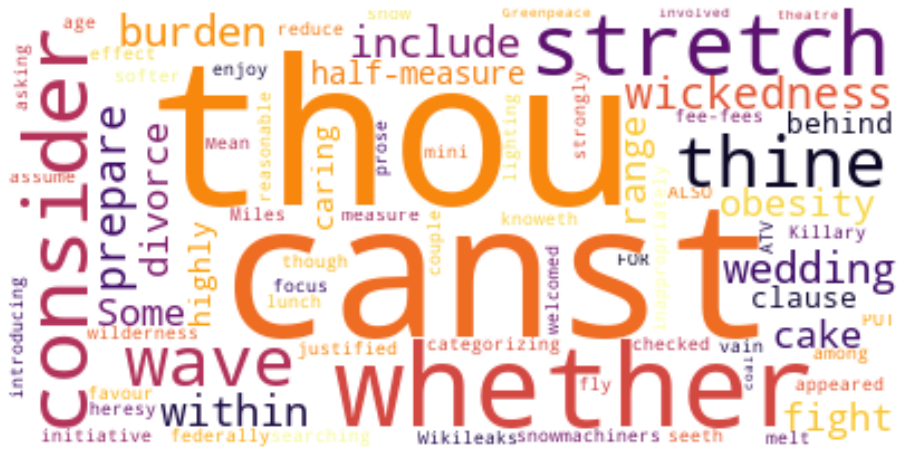}\label{fig:wordcloud-clean-gri}}}%
    \caption{Global explanation of toxic comment classification with LSTM.}
    \label{fig:wordcloud-lstm}%
\end{figure*}

\begin{figure*}[h!]%
    \centering
    \subfloat[GAI \textit{Positive} class.]{{
    \includegraphics[width=0.4\textwidth]{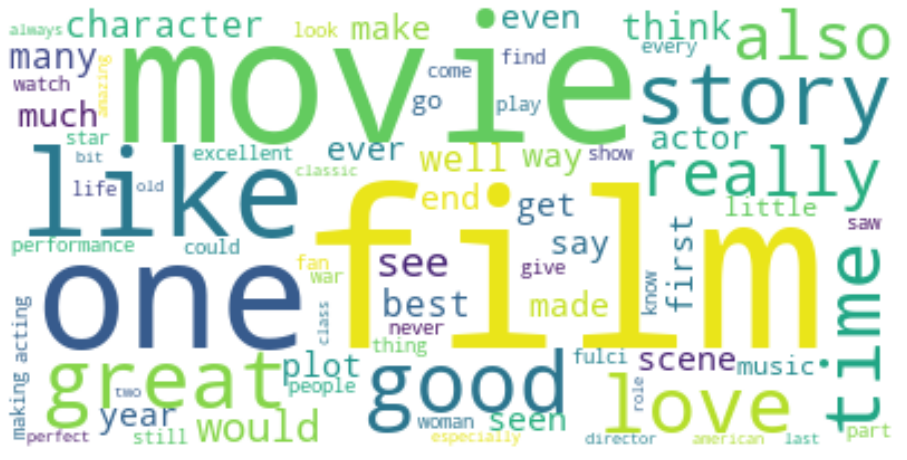}\label{fig:wordcloud-positive-gai}}}%
    \qquad
    \subfloat[GAI \textit{Negative} class.]{{
    \includegraphics[width=0.4\textwidth]{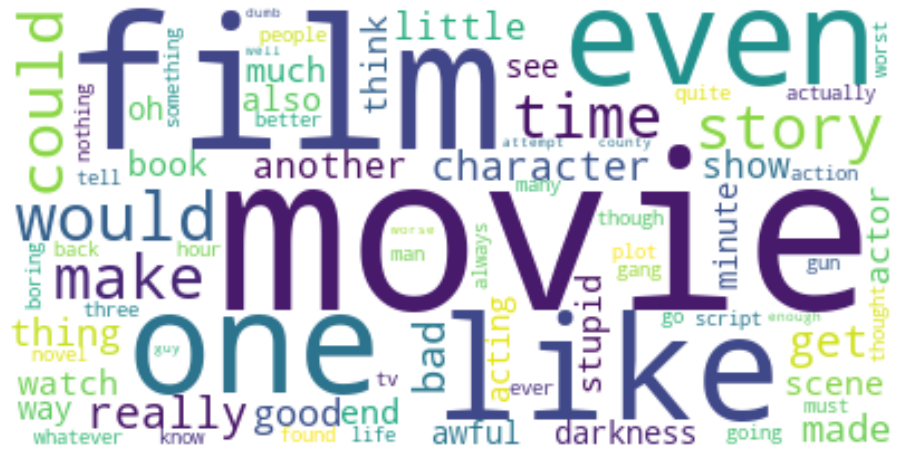}\label{fig:wordcloud-negative-gai}}}%
    
    \subfloat[GRI \textit{Positive} class.]{{
    \includegraphics[width=0.4\textwidth]{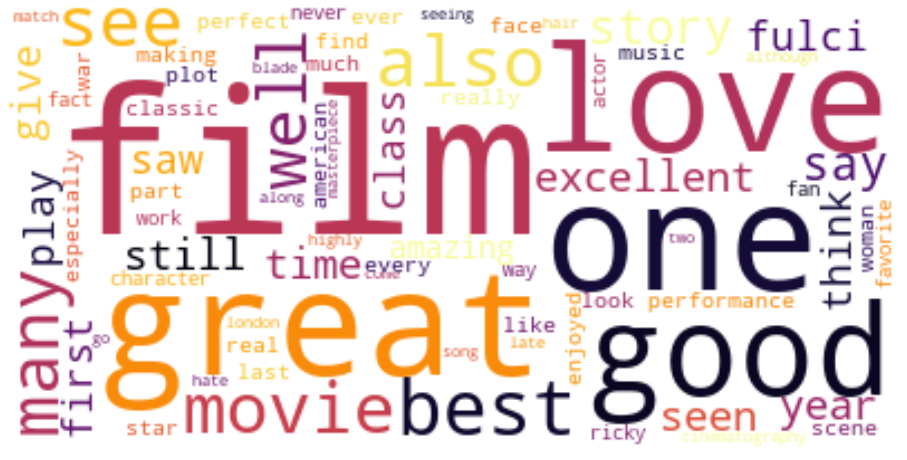}\label{fig:wordcloud-positive-gri}}}%
    \qquad
    \subfloat[GRI \textit{Negative} class.]{{
    \includegraphics[width=0.4\textwidth]{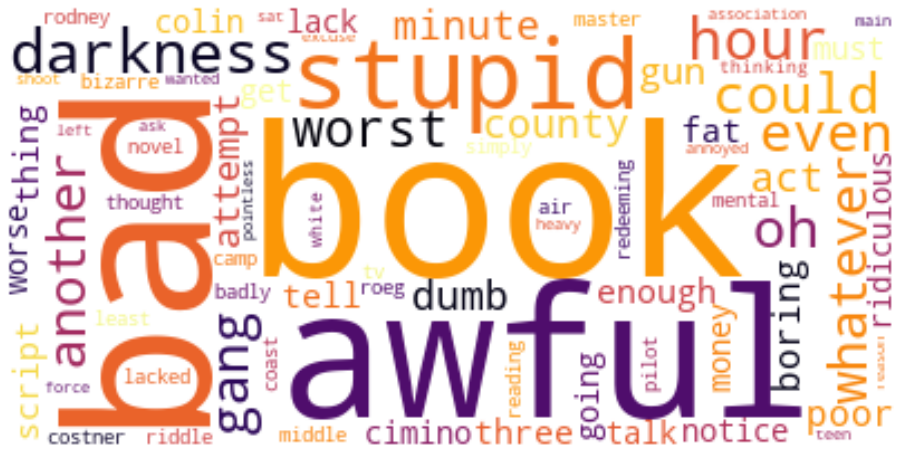}\label{fig:wordcloud-negative-gri}}}%
    \caption{ Global explanation of sentiment analysis with BERT.}
    \label{fig:wordcloud-bert}%
\end{figure*}

\vspace{0.2cm}
\noindent
\textbf{Use case 1.} For the toxic comment classification,
Figure~\ref{fig:wordcloud-lstm} shows the $GAI$ and $GRI$ scores for each influential word under the form of word-clouds for the classes \textit{Toxic} (Figure~\ref{fig:wordcloud-toxic-gai} and \ref{fig:wordcloud-toxic-gri}) and \textit{Clean} (Figure~\ref{fig:wordcloud-clean-gai} and \ref{fig:wordcloud-clean-gri}), respectively.
The font size of words is proportional to the $GAI$ or $GRI$ scores obtained for each class separately. The proportion of the font size is relative only to the single word-cloud (i.e., two words with the same size in different word-clouds do not necessarily have the same score, while two words with the same size in the same word-cloud have almost the same score).
The $GAI$ word-clouds (Figures~\ref{fig:wordcloud-toxic-gai} and \ref{fig:wordcloud-clean-gai}) show that the two classes are influenced by a non-overlapping set of words.
This confirms that the model learned that if a word is attributable to toxic language in some context, it is unlikely to be associated to clean language in others.
Toxic comments are identified by terms that are strongly related to toxic language, discrimination, or racism.
Instead, there is no specific pattern of words that identifies clean comments.
Just few concepts like \texttt{people} have an inter-class influence.

Then, the $GRI$ word-clouds (Figures~\ref{fig:wordcloud-toxic-gri} and \ref{fig:wordcloud-clean-gri}) determine which are the more differentiating concepts between the two classes, among those selected by the model.
The $GRI$ word-cloud highlights even more the impact of words like \texttt{stupid}, \texttt{idiot} and \texttt{ignorant}, but also terms related to minorities and genders like \texttt{woman}, \texttt{black}, \texttt{white}, \texttt{gay}, meaning that the model has learned to recognize racists or sexists comments when these terms are present. 
Also, the presence of specific politician family names, anonymized as \texttt{Politician1}, \texttt{Politician2}, etc., highlight that those people names are related to toxic comments. 
These results demonstrate that a black-box model, if not carefully trained, can learn from sensible content including prejudices and various forms of bias that should be avoided in critical contexts.
Finally, associating a specific person's family name to a class also raises ethical issues.

\vspace{0.2cm}
\noindent
\textbf{Use case 2.} 
Analyzing the prediction-local explanations produced for the 400 input texts in the \textit{sentiment analysis} use case, it is possible to extract global insights regarding the fine-tuned BERT model.
Figure~\ref{fig:wordcloud-bert} shows the $GAI$ and $GRI$ word-clouds for the \textit{Positive} (Figure~\ref{fig:wordcloud-positive-gai} and \ref{fig:wordcloud-positive-gri}) and \textit{Negative} (Figure~\ref{fig:wordcloud-negative-gai} and \ref{fig:wordcloud-negative-gri}) class labels.

Differently than the previous example, the $GAI$ word-clouds for the \textit{Positive} (Figure~\ref{fig:wordcloud-positive-gai}) and the \textit{Negative} (Figure~\ref{fig:wordcloud-negative-gai}) class labels show that several words like \texttt{story}, \texttt{movie}, \texttt{film}, \texttt{like} are impacting on both classes.
This means that the model exploits overlapping concepts that do not express directly a sentiment but that, if considered together in their context, can be associated with words that express the mood of the writer (e.g. \texttt{This film is not as good as expected}).

The $GRI$ word-cloud for the \textit{Positive} class (Figure~\ref{fig:wordcloud-positive-gri}) shows that words like \texttt{movie} and \texttt{film} are still very relevant for it, while they do not appear anymore for the \textit{Negative} class (Figure~\ref{fig:wordcloud-negative-gri}) that is now highly characterized by the concept of \texttt{book}.
Exploring the dataset, we noticed that movies inspired by books are used to be associated with negative comments, as typically the original book is more detailed or slightly different, and thus this can be considered a form of bias that the model has learned, in the sense that a movie evaluation might not be based on its comparison with a book. 
However, the $GRI$ shows also that most of the influential words for positive input texts are concepts strictly related to positive sentiments like \texttt{good}, \texttt{great}, \texttt{best}, \texttt{love}. 
Similarly, the negative sentiment is associated to words like \texttt{worst}, \texttt{bad}, \texttt{awful}. For these concepts, the model behaves as expected.

Thanks to the model-global explanation process the user can better understand how the predictive model is taking its decisions, identifying the presence of prejudice and/or bias, and allowing to decide if and which corrective actions have to be taken to make the decision-making process more reliable.
\section{Conclusion and future research directions}
\label{sec:conclusion}

This paper proposed \XXX, a new engine able to provide both prediction-local and model-global interpretable explanations in the context of NLP analytics tasks that exploit black-box deep-learning models.
\XXX's experimental assessment includes two different NLP tasks, i.e., a sentiment analysis task and a toxic document classification, performed through state-of-the-art techniques: a fine-tuned BERT model and a custom LSTM model.

Results showed that \XXX\ can 
(i) identify specific features of the textual input data that are predominantly influencing the model's predictions, 
(ii) highlight such features to the end-user, and 
(iii) quantify their impact through novel indexes.
The proposed explanations enable end-users to decide whether a specific local prediction made by a black-box model is reliable, and to evaluate the general behavior of the global model across predictions.
Besides being useful to general-purpose end users, explanations provided by \XXX\ are especially useful for data scientists, artificial intelligence and machine learning experts in need of understanding the behavior of their models, since the extracted features, both textual and numeric, are an efficient way to harness the complex knowledge learned by the models themselves.
Future research directions include: 
(a) Further investigating new strategies for the perturbation of the input features, such as the substitution perturbation;
(b) integrating \XXX\ in a real-life setting to measure the effectiveness of the proposed textual explanations by human validation, interviewing both expert and non-expert users; 
(c) generalizing \XXX\ to include new analytics goals, such as fine-tuned strategies and concept-drift detection.

\bibliography{elsarticle-main.bib}

\end{document}


\begin{frontmatter}

\title{Elsevier \LaTeX\ template\tnoteref{mytitlenote}}
\tnotetext[mytitlenote]{Fully documented templates are available in the elsarticle package on \href{http://www.ctan.org/tex-archive/macros/latex/contrib/elsarticle}{CTAN}.}

\author{Elsevier\fnref{myfootnote}}
\address{Radarweg 29, Amsterdam}
\fntext[myfootnote]{Since 1880.}

\author[mymainaddress,mysecondaryaddress]{Elsevier Inc}
\ead[url]{www.elsevier.com}

\author[mysecondaryaddress]{Global Customer Service\corref{mycorrespondingauthor}}
\cortext[mycorrespondingauthor]{Corresponding author}
\ead{support@elsevier.com}

\address[mymainaddress]{1600 John F Kennedy Boulevard, Philadelphia}
\address[mysecondaryaddress]{360 Park Avenue South, New York}

\begin{abstract}
This template helps you to create a properly formatted \LaTeX\ manuscript.
\end{abstract}

\begin{keyword}
\texttt{elsarticle.cls}\sep \LaTeX\sep Elsevier \sep template
\MSC[2010] 00-01\sep  99-00
\end{keyword}

\end{frontmatter}

\linenumbers

\begin{figure}[!ht]
    \centering
    \includegraphics[width=1\columnwidth]{img/Algorithm overview_4.png}
    \caption{\XXX\ overview}
    \label{fig:architecture-overview}
\end{figure}

\begin{figure}
\centering
\includegraphics[width=0.4\textwidth]{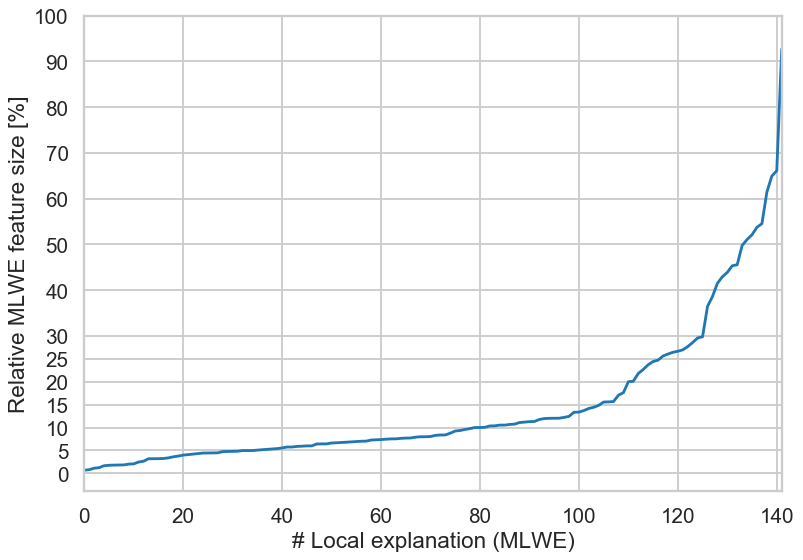}
\caption{Relative size of the features extracted with MLWE strategy (i.e. size of the most influential clusters of words) for all the informative local explanations. The relative size is calculated, for each local explanation, as the number of tokens in the cluster over the number of tokens in the input text.}
\label{fig:mlwe-rel-feat-size}
\end{figure}

\begin{figure}
    \centering
    \subfloat[Original text]{
        \label{fig:ex-3-original}
        \fbox{\begin{minipage}{1\columnwidth}
        \scriptsize
        \fontfamily{cmtt}\selectfont
        They are nuts that have the Liberal government by the short and curlies, the good news is we see them a d they are being rejected on a global stage! It can't happen fast enough, the Marxist pigs are a freak show better suited for a circus.
        \end{minipage}}
    }\\
    \subfloat[\texttt{EXP1:} Noun - POS feature extraction]{
        \label{fig:ex-3-noun}
        \fbox{\begin{minipage}{1\columnwidth}
        \scriptsize
        \fontfamily{cmtt}\selectfont
        They are \textcolor{ansi-red-intense}{nuts} that have the Liberal \textcolor{ansi-red-intense}{government} by the short and \textcolor{ansi-red-intense}{curlies}, the good \textcolor{ansi-red-intense}{news} is we see them a \textcolor{ansi-red-intense}{d} they are being rejected on a global \textcolor{ansi-red-intense}{stage}! It can't happen fast enough, the \textcolor{ansi-red-intense}{Marxist pigs} are a freak \textcolor{ansi-red-intense}{show} better suited for a \textcolor{ansi-red-intense}{circus}.
        \end{minipage}}
    }\\
   
    \subfloat[\texttt{EXP2:} Multi-layer word embedding feature extraction]{
        \label{fig:ex-3-mlwe}
        \fbox{\begin{minipage}{1\columnwidth}
        \scriptsize
        \fontfamily{cmtt}\selectfont
        They are \textcolor{ansi-red-intense}{nuts} that have the Liberal government by the short and curlies, the good news is we see them a d they are being rejected on a global stage! It can't happen fast enough, the Marxist \textcolor{ansi-red-intense}{pigs} are a \textcolor{ansi-red-intense}{freak} show better \textcolor{ansi-red-intense}{suited} for a \textcolor{ansi-red-intense}{circus}.
        \end{minipage}}
    }
    
    \caption{Examples of \textit{textual explanation} report for the input in Figure~\ref{fig:ex-3-original} originally labeled by custom LSTM model as \textit{Toxic} with a probability of 0.96. The most relevant features are highlighted in red.}
    \label{fig:ex-3-textual-explanation}
    
    \vspace{1em}
    
    \begin{minipage}{\columnwidth}
        \centering
        \scriptsize
        \begin{tabular}{|c|c|c|c|c|}
            \hline
            Explanation & Feature $f$ & $L_o$ & $L_f$  & $nPIR_f(N)$ \\ \hline
             EXP1 & POS-Noun & T & C & 0.608 \\ \hline
             EXP2 & MLWE & T & C & 0.999 \\ \hline

        \end{tabular}
        \captionof{table}{Quantitative explanation for example in Figure~\ref{fig:ex-3-textual-explanation}. T is the Toxic label, C is the Clean  label.}
        \label{tab:ex-3-numerical-explanation}
    \end{minipage}
\end{figure}

In the third example, Figure~\ref{fig:ex-3-textual-explanation}, the custom LSTM model has been exploited analyzing the decision-making process that brought it to classify the input document in Figure~\ref{fig:ex-3-original} as \textit{Toxic}. The most influential features extracted by \XXX\ are showed in Figures~\ref{fig:ex-3-noun} and \ref{fig:ex-3-mlwe}, while the \textit{quantitative explanation} is reported in Table~\ref{tab:ex-3-numerical-explanation}. An analysis process of the explanation reports similar to the one performed in the previous examples can be followed even in this case.
The POS analysis identifies the \textit{Nouns} as more influential part-of-speech for the prediction of the \textit{Toxic} class label with a nPIR score of 0.608 (EXP1). Moreover, the MLWE feature analysis, finds the cluster of words \{{\fontfamily{cmtt}\selectfont nuts, pigs, freak, suited, circus}\} as a very positively important feature for the prediction of class \textit{Toxic} with an nPIR score of 0.999 (EXP2). As can be noticed, the two features are quite similar, but the MLWE can find a tiny number of words that together cause the prediction of the class label. 
Moreover, the MLWE features appear to be significant for the performed prediction, enforcing the trustfulness of the performed prediction. 
Thus, thanks to the provided explanations it is straightforward to confirm the correctness of the decision-making process performed by the LSTM model for this specific input document, from the moment that \XXX\ identified as influential specific words that are usually correlated to a toxic language.

\section*{References}

\bibliography{mybibfile}